%% file: arxiv.tex
\documentclass{article}
\usepackage{ijcai21}

\usepackage{times}

\usepackage{soul}
\usepackage{array}
\usepackage{ragged2e}
\usepackage{amssymb}
\usepackage{float}
\usepackage{dcolumn}
\usepackage{multirow}
\usepackage{url}
\usepackage[hidelinks]{hyperref}
\usepackage[utf8]{inputenc}
\usepackage[small]{caption}
\usepackage{graphicx}
\usepackage{amsmath}
\usepackage{booktabs}
\usepackage{amsmath}
\usepackage{amssymb}
\usepackage{array}
\usepackage{graphicx}
\usepackage[usenames, dvipsnames]{color}
\usepackage{rotating}
\usepackage{multirow}
\usepackage{pgfplots}
\usepackage{algorithm}
\usepackage{algorithmicx}
\usepackage[noend]{algpseudocode}
\usepackage{hyperref}
\usepackage[margin=4pt,font=footnotesize,labelfont=bf,labelsep=endash,tableposition=top]{caption}
\usepackage{amsmath}
\usepackage{cases}
\usepackage{times}
\usepackage{subcaption}
\usepackage{epsfig}
\usepackage{graphicx}
\usepackage{amsmath}
\usepackage{amssymb}
\usepackage{multirow}
\usepackage{multicol}
\usepackage{algorithmicx}
\usepackage{enumitem}
\usepackage{arydshln}
\usepackage{bm}

\title{Segmenting Transparent Object in the Wild with Transformer}

\author{
    Enze Xie$^{1}$\ ,
    Wenjia Wang$^{2}$,
    Wenhai Wang$^{3}$, 
    Peize Sun$^{1}$, \\
    Hang Xu$^1$,
    Ding Liang$^2$, 
    Ping Luo$^1$ \\
    \affiliations 
    $^1$The University of Hong Kong~~~
    $^2$Sensetime Research~~~
    $^3$Nanjing University
    \\
}

\begin{document}

\maketitle

\begin{abstract}
This work presents a new fine-grained transparent object segmentation dataset, termed Trans10K-v2, extending Trans10K-v1, the first large-scale transparent object segmentation dataset. 
Unlike Trans10K-v1 that only has two limited categories, our new dataset has several appealing benefits. (1) It has 11 fine-grained categories of transparent objects, commonly occurring in the human domestic environment, making it more practical for real-world application. 
(2) Trans10K-v2 brings more challenges for the current advanced segmentation methods than its former version.
Furthermore, a novel transformer-based segmentation pipeline termed Trans2Seg is proposed.
Firstly, the transformer encoder of Trans2Seg provides the global receptive field in contrast to CNN's local receptive field, which shows excellent advantages over pure CNN architectures.
Secondly, by formulating semantic segmentation as a problem of dictionary look-up, we design a set of learnable prototypes as the query of Trans2Seg's transformer decoder, where each prototype learns the statistics of one category in the whole dataset.
We benchmark more than 20 recent semantic segmentation methods, demonstrating that Trans2Seg significantly outperforms all the CNN-based methods, showing the proposed algorithm's potential ability to solve transparent object segmentation.
Code is available in \href{https://github.com/xieenze/Trans2Seg}{\color{blue}{\tt github.com/xieenze/Trans2Seg}}.
\end{abstract}

\section{Introduction}
Modern robots, mainly mobile robots and mechanical manipulators, would benefit a lot from the efficient perception of the transparent objects in residential environments since the environments vary drastically. The increasing utilization of glass wall and transparent door in the building interior and the glass cups and bottles in residential rooms has resulted in the wrong detection in various range sensors.
In robotic research, most systems perceive the environment by multi-data sensor fusion via sonars or lidars. The sensors are relatively consistent in detecting opaque objects but are still affected by the scan mismatching due to transparent objects. The unique feature of reflection, refraction, and light projection from the transparent objects may confuse the sensors. \textit{Thus a reliable vision-based method, which is much cheaper and more robust than high-precision sensors, would be efficient.}

\begin{figure}[t]
  \centering
    \begin{subfigure}{0.47\textwidth}
      \centering   
      \includegraphics[width=1\linewidth]{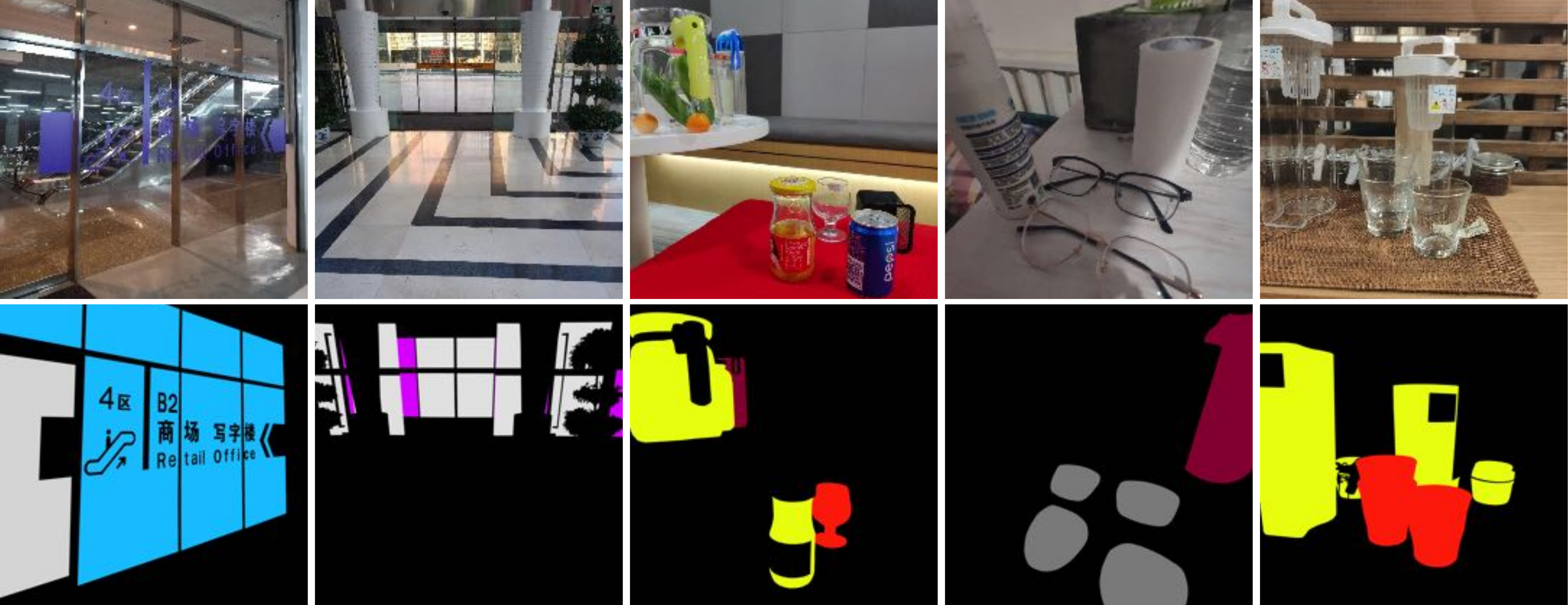}
        \caption{Selected images and corresponding high-quality masks.}
        \label{fig:sub1}
    \end{subfigure}   
    \begin{subfigure}{0.47\textwidth}
      \centering   
      \includegraphics[width=1\linewidth]{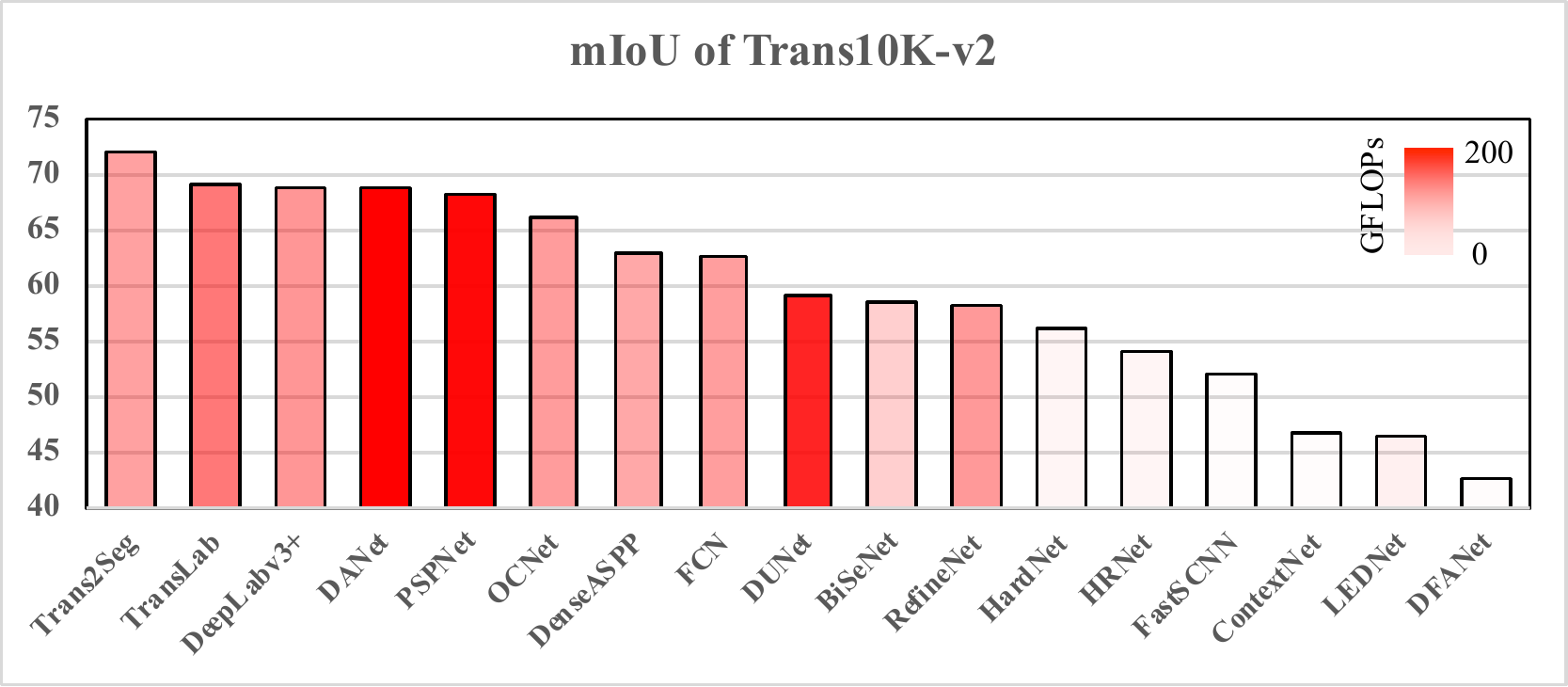}
        \caption{Performance comparison on Trans10K-v2.}
        \label{fig:exp_stat}
    \end{subfigure}
    \vspace{-5pt}
\caption{\label{fig:total1}
(a) shows the high diversity of our dataset and high-quality annotations.  
(b) is \textbf{Comparisons} between Trans2Seg and other CNN-based semantic segmentation methods. All methods are trained on Trans10K-v2 with same epochs. mIoU is chosen as the metric. Deeper color bar indicates methods with larger FLOPS. Our Trans2Seg significantly surpasses other methods with lower flops.
}
\vspace{-10pt}
\end{figure}

Although some transparent objects dataset~\cite{transcut,tomnet,GDNet} were proposed, there are some obvious problems.
(1) Limited dataset scale. These datasets often have less than 1K images captured from the real-world and less than 10 unique objects.
(2) Poor diversity. The scene of these datasets is monotonous.
(3) Fewer classes. All these datasets have only two classes, background and transparent objects. They lack fine-grained categories, which limited their practicality.
Recently, \cite{translab} proposed a large-scale and high-diversity dataset termed Trans10K, which divide transparent objects as `Things' and `Stuff'. The dataset is high diversity, but it also lacks fine-grained transparent categories.

In this paper, we proposes a fine-grained transparent object segmentation dataset termed Trans10K-v2 with more elaborately defined categories. The images are inherit from Trans10K-v1~\cite{translab}.
We annotate the 10428 images with 11 fine-grained categories: shelf, jar, freezer, window, glass door, eyeglass, cup, glass wall, glass bowl, water bottle, storage box. 
In Trans10K-v1, transparent~\textbf{things} are defined to be grabbed by the manipulators and \textbf{stuff} are for robot navigation. Though two basic categories can partially help robots to interact with transparent objects, the provided fine-grained classes in Trans10K-v2 can provide more. We analyze these objects' functions and how robots interact with them in appendix.

Based on this challenging dataset, we design Trans2Seg, introducing Transformer into segmentation pipeline for its encoder-decoder architecture. First, the transformer encoder provides a global receptive field via self-attention. Larger receptive field is essential for segmenting transparent objects because transparent objects often share similar textures and context with its surroundings. 
Second, the decoder stacks successive layers to interact query embedding with transformer encoder output. To facilitate the robustness of transparent objects, we carefully design a set of learnable class prototype embeddings as the query for transformer decoder and the key is the feature map from the transformer encoder. Compared with convolutional paradigm, where the class prototypes is the fixed parameters of convolution kernel weight, our design provides a dynamic and context-aware implementation. 
As shown in Figure.~\ref{fig:exp_stat}, we train and evaluate 20 existing representative segmentation methods on Trans10K-v2, and found that simply applying previous methods to this task is far from sufficient. By successfully introducing Transformer into this task, our Trans2Seg significantly surpasses the best TransLab~\cite{translab} by a large margin~(72.1 \textit{vs.} 69.0 on mIoU).

In summary, our main contributions are three-fold:
\begin{itemize}
\item We propose the largest glass segmentation dataset (Trans10K-v2) with 11 fine-grained glass image categories with a diverse scenario and high resolution. All the images are elaborately annotated with fine-shaped masks and function-oriented categories.
\item
We introduce a new transformer-based network for transparent object segmentation with transformer encoder-decoder architecture. Our method provides a global receptive field and is more dynamic in mask prediction, which shows excellent advantages.
\item
We evaluate more than 20 semantic segmentation methods on Trans10K-v2, and our Trans2Seg significantly outperforms these methods. Moreover, we show this task is largely unsolved. Thus more research is needed.
\end{itemize}

\section{Related Work}
\textbf{Semantic Segmentation.}
In deep learning era, convolutional neural network (CNN) puts forwards the development of semantic segmentation in various datasets, such as ADE20K, CityScapes and PASCAL VOC. One of the pioneer works approaches, FCN~\cite{fcn}, transfers semantic segmentation into an end-to-end fully convolutional classification network. For improving the performance, especially around object boundaries,~\cite{deeplab,lin2016efficient,zheng2015conditional} propose to use structured prediction module, conditional random fields (CRFs)~\cite{crf}, to refine network output. Dramatic improvements in performance and inference speed have been driven by aggregating features at multiples scales, for example, PSPNet~\cite{pspnet} and DeepLab~\cite{deeplab,deeplab2}, and propagating structured information across intermediate CNN representations~\cite{gadde2016superpixel,liu2017learning,nonlocal}.

\textbf{Transparent Object Datasets.}
\cite{transcut} introduces TransCut dataset which only contain 49 images of 7 unique objects. To generate the segmentation result, \cite{transcut} optimized an energy function based on LF-linearity which also need to utilize the light-field cameras.
\cite{tomnet} proposed TOM-Net. It contains 876 real images and 178K synthetic images which are generated by POV-Ray. However, only 4 unique objects are used in synthesizing the training data.
Recnetly, \cite{translab} introduce a first large-scale real-world transparent object segmentation dataset, termed Trans10K. It has 10K+ images. However, there are two categories in this dataset, which limits its practical use. In this work, our Trans10K-v2 inherited the data and annotates 11 fine-grained categories.

\begin{figure*}[t]
    \centering
    \scalebox{0.17}{\includegraphics{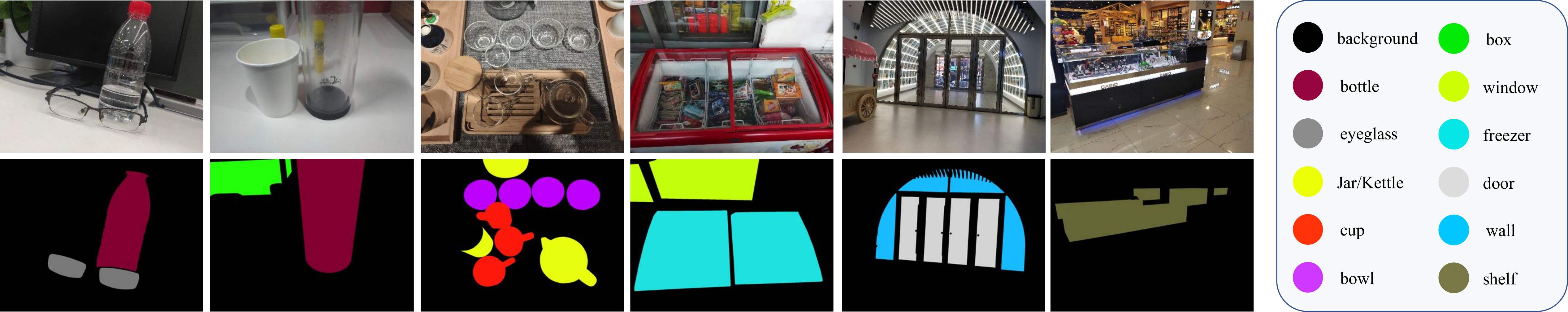}}
    \caption{\textbf{Images in Trans10K-v2 dataset are carefully annotated with high quality.} The first row shows sample images and the second shows the segmentation masks. 
    The color scheme which encodes the object categories are listed on the right of the figure. Zoom in for best view.}
    \label{fig:visualize}
    \vspace{-10pt}
\end{figure*}

\begin{table*}[t]
\centering
 \scalebox{1.1}
 {\small\input{tables/stat2}}
\vspace{2mm}
\caption{\textbf{Statistic information of Translabv2.} `CMCC' denotes Mean Connected Components of each category. `image num' denotes the image number. `pixel ratio' is the pixel number of a certain category accounts in all the pixels of transparent objects in Trans10K-v2.}
\label{tab:stat}
\vspace{-20pt}
\end{table*}

\textbf{Transformer in Vision Tasks.}
Transformer~\cite{vaswani2017attention} has been successfully applied in both high-level vision and low-level vision~\cite{han2020survey}. In  ViT~\cite{dosovitskiy2020image}, Transformer is directly applied to sequences of image patches to complete image classification. In object detection areas~\cite{DETR,deformdetr}, DETR reasons about the relations of the object queries and the global image context via Transformer and outputs the final set of predictions in parallel without non-maximum suppression(NMS) procedures and anchor generation.
SETR~\cite{zheng2020setr} views semantic segmentation from a sequence-to-sequence perspective with Transformer.
IPT~\cite{chen2020pre} applies Transformer model to low-level computer vision task, such as denoising, super-resolution and deraining. In video processing, Transformer has received significantly growing attention.  VisTR~\cite{wang2020end} accomplishes instance sequence segmentation by Transformer. Multiple-object tracking~\cite{transtrack,trackformer} employs Transformers to decode object queries and feature queries of the previous frame into bounding boxes of the current frame, and merged by Hungarian Algorithm or NMS.

\section{Trans10K-v2 Dataset}
\textbf{Dataset Introduction.}
Our Trans10K-v2 dataset is based on Trans10K dataset~\cite{translab}. 
Following Trans10K, we use 5000, 1000 and 4428 images in training, validation and testing respectively. The distribution of the images is abundant in occlusion, spatial scales, perspective distortion. We further annotate the images with more fine-grained categories due to the functional usages of different objects.
Trans10K-v2 dataset contains 10,428 images, with two main categories and 11 fine-grained categories: (1) Transparent \textbf{Things} containing \textbf{cup}, \textbf{bottle}, \textbf{jar}, \textbf{bowl} and \textbf{eyeglass}. (2) Transparent \textbf{Stuﬀ} containing \textbf{windows}, \textbf{shelf}, \textbf{box}, \textbf{freezer}, \textbf{glass walls} and \textbf{glass doors}. 
In respect to ﬁne-grained categories and high diversity, Trans10K-v2 is very challenging, and have promising potential in both computer vision and robotic researches.

\textbf{Annotation Principle.}
The transparent objects are manually labeled by expert annotators with professional labeling tool. The annotators were asked to provide more than 100 points when they trace the boundaries of each transparent object, which ensures the high-quality outline of the mask shapes. The way of annotation is mostly the same with semantic segmentation datasets such as ADE20K. We set the background with 0, and the 11 categories from 1 to 11. We also provide the scene environment of each image locates at. The annotators are asked to strictly following principles when they label the images: (\uppercase\expandafter{\romannumeral1}) Only highly transparent pixels are annotated as masks, other semi-transparent and non-transparent pixels are ignored. 
Highly transparent objects no matter made of glass, plastics or crystals should also be annotated. (\uppercase\expandafter{\romannumeral2}) When occluded by opaque objects, the pixels will be cropped from the masks. (\uppercase\expandafter{\romannumeral3}) 
The setting of all 11 fine-grained categories are elaborately observed and induced from the point of function. 
We analyze firstly how the robots need to deal with the transparent objects as avoiding or grasping or manipulating, then categorize the objects similar in shape and function into a fine-grained category.
The detailed principle of how we categorize the objects is listed in appendix.

\textbf{Dataset Statistics.}
The statistic information of 
CMCC, imaga number, pixel proportion are listed in Table~\ref{tab:stat} in detail. From Table\ref{tab:stat}, the sum of all the image numbers is larger than 10428 since some image has multiple category of objects. CMCC denotes Mean Connected Components of each category. It is caculated by dividing the connected components number of a certain category by the image number. The number of connected components are counted by the boundary of the masks. It represents the complexity of the transparent objects.

\textbf{Evaluation Metrics.}
Results are reported in three metrics that are widely used in semantic segmentation to benchmark the performance of fine-grained transparent object segmentation. 
\textbf{(1) Pixel Accuracy} indicates the proportion of correctly classified pixels.
\textbf{(2) Mean IoU} indicates mean intersection over union.
\textbf{(3) Category IoU} indicates the intersection over union of each category.

\section{Method}

\subsection{Overall Pipeline}
\begin{figure*}[t]
    \centering
    \scalebox{0.58}{\includegraphics{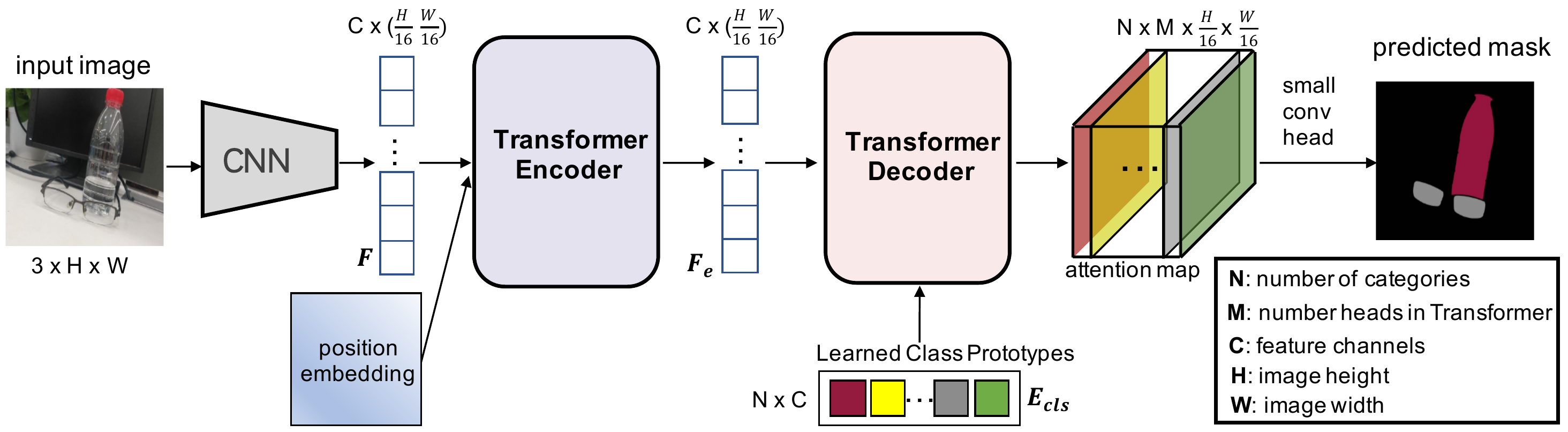}}
    \caption{\textbf{The whole pipeline of our hybrid CNN-Transformer architecture.}
    First, the input image is fed to CNN to extract features $F$. 
    Second, for transformer encoder, the features and position embedding are flatten and fed to transformer for self-attention, and output feature($F_e$) from transformer encoder. 
    Third, for transformer decoder, we specifically \textbf{define a set of learnable class prototype embeddings($E_{cls}$) as query, $F_e$ as key}, and calculate the attention map with $E_{cls}$ and $F_e$. Each class prototype embedding corresponds to a category of final prediction. We also add a small conv head to fuse attention map and Res2 feature from CNN backbone. Details of transformer decoder and small conv head refer to Figure~\ref{fig:decoder}.
    Finally, we can get the predict results by doing pixel-wise argmax on the attention map.
    For example, in this figure, the segmentation mask of two categories~(\textbf{\textcolor[RGB]{137,0,56}{Bottle}} and \textbf{\textcolor[RGB]{129,129,129}{Eyeglass}}) corresponds to two class prototypes with same colors.
    }
    \label{fig:Figure3}
    \vspace{-10pt}
\end{figure*}

The overall Trans2Seg architecture contains a CNN backbone, an encoder-decoder transformer, and a small convolutional head, as shown in Figure~\ref{fig:Figure3}. For an input image of $(H, W, 3)$, 
\begin{itemize}
    \item The CNN backbone generates image feature map of $(\frac{H}{16}, \frac{W}{16}, C)$.
    \item The encoder takes in the summation of flattened feature of $(\frac{H}{16}\frac{W}{16},C)$ and positional embedding of $(\frac{H}{16}\frac{W}{16},C)$, and outputs encoded feature of $(\frac{H}{16}\frac{W}{16},C)$.
    \item The decoder interacts the learned class prototypes of $(N,C)$ with encoded feature, and generates attention map of $(N,M,\frac{H}{16} \frac{W}{16})$, where $N$ is number of categories, $M$ is number of heads in multi-head attention.
    \item The small convolutional head up-samples the attention map to $(N, M, \frac{H}{4}, \frac{W}{4})$,  fuses it with high-resolution feature map Res2 and outputs attention map of $(N, \frac{H}{4}, \frac{W}{4})$. 
\end{itemize}
The final segmentation is obtained by pixel-wise argmax operation on the output attention map.

\begin{figure}[H]
    \centering
    \scalebox{0.40}{\includegraphics{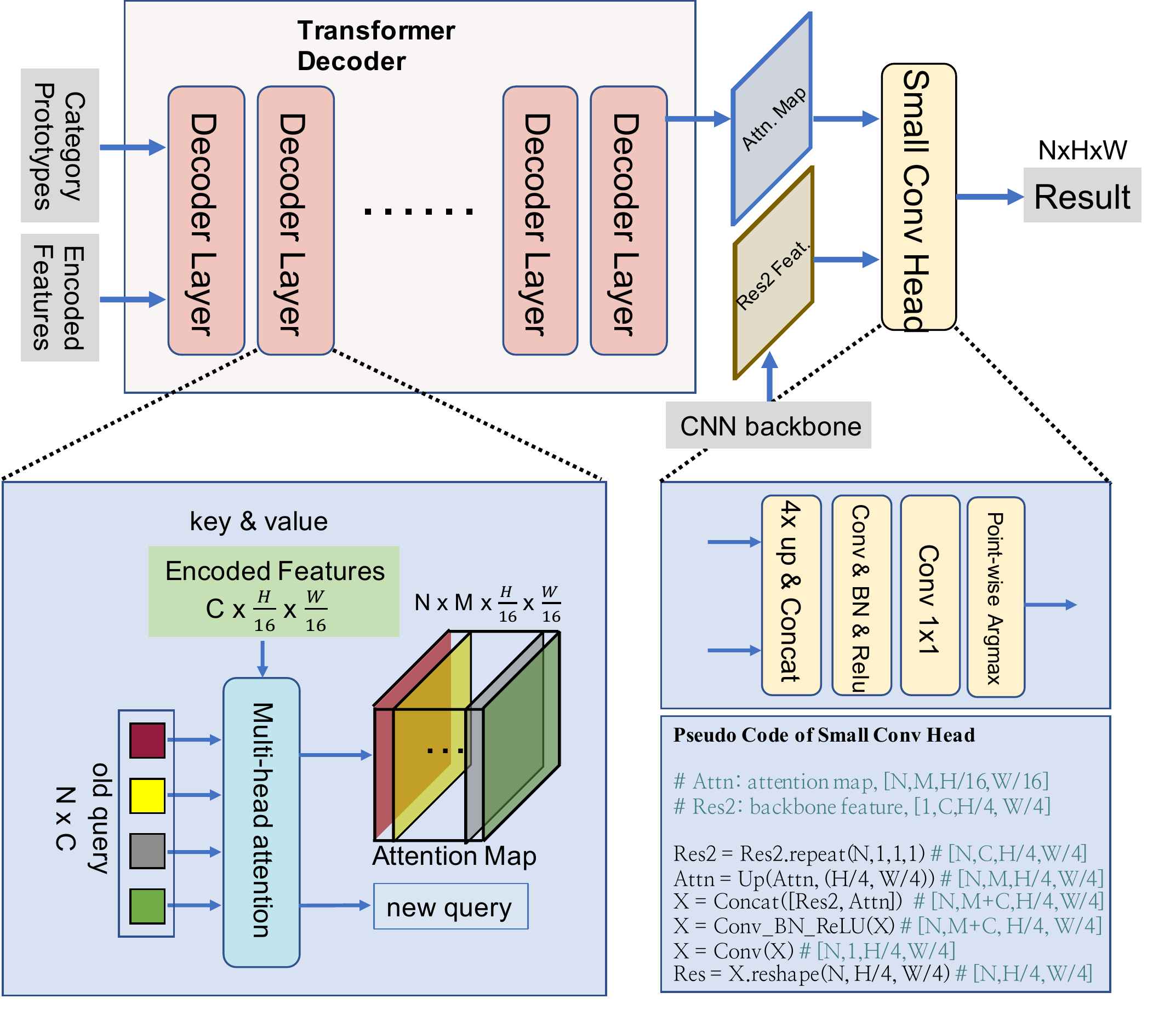}}
    \caption{\textbf{Detail of Transformer Decoder and small conv head.} Input: The learnable category prototypes as query, features from transformer encoder as key and value. The inputs are fed to transformer decoder, which consists of several decoder layers. The attention map from last decoder layer and the Res2 feature from CNN backbone are combined and fed to a small conv head to get final prediction result. We also provide the \textit{Pseudo Code of small conv head} for better understanding. 
    }
    \label{fig:decoder}
\end{figure}

\subsection{Encoder}
The Transformer encoder takes a sequence as input, so the spatial dimensions of the feature map $(\frac{H}{16}, \frac{W}{16}, C)$ is flattened into one dimension$(\frac{H}{16}\frac{W}{16},C)$. To compensate missing spatial dimensions, positional embedding~\cite{posencoding} is supplemented to one dimension feature to provide information about the relative or absolute position of the feature in the sequence. The positional embedding has the same dimension $(\frac{H}{16}\frac{W}{16},C)$ with the flattened feature. The encoder is composed of stacked encoder layers, each of which consists of a multi-head self-attention module and a feed forward network~\cite{vaswani2017attention}.

\subsection{Decoder}
The Transformer decoder takes input a set of learnable class prototype embeddings as query, denoted by $E_{cls}$, the encoded feature as key and value, denoted by $F_e$, and output the attention map followed by Small Conv Head to obtain final segmentation result, as shown in Figure~\ref{fig:decoder}.

The class prototype embeddings are learned category prototypes, updated iteratively by a series of decoder layers through multi-head attention mechanisms. We denoted iterative update rule by $\bigodot$, then the class prototype in each decoder layer is:
\begin{equation}
    \rm{E_{cls}^s} = \bigodot_{i=0,..,s-1}\rm{softmax}(E_{cls}^iF_e)F_e  
\end{equation}

In the final decoder layer, the attention map is extracted out to into small conv head: 
\begin{equation}
     \rm{attention \  map} = E_{cls}^sF_e   
\end{equation}

The pseudo code of small conv head is shown in shown in Figure~\ref{fig:decoder}. The attention map from Transformer decode is the shape of $(N,M,\frac{H}{16} \frac{W}{16})$, where $N$ is number of categories, $M$ is number of heads in multi-head attention. It is up-sampled to $(N, M, \frac{H}{4}, \frac{W}{4})$, then fused with high-resolution feature map Res2 in the second dimension to $(N, M+C, \frac{H}{4}, \frac{W}{4})$, and finally transformed into output attention map of $(N, \frac{H}{4}, \frac{W}{4})$. The final segmentation is obtained by pixel-wise argmax operation on the output attention map.

\subsection{Discussion}
The most related work with Trans2Seg is SETR and DETR~\cite{zheng2020setr,DETR}. In this section we discuss the relations and differences in details.

\textbf{SETR}. Trans2Seg and SETR are both segmentation pipelines. Their key difference is reflected in the design of the decoder. In SETR, the decoder is simple several convolutional layers, which is similar with most previous methods. However, the decoder of Trans2Seg is also transformer, which fully utilize the advantages of attention mechanism in semantic segmentation.

\textbf{DETR}. Trans2Seg and DETR share similar components in the pipeline, including CNN backbone, Transformer encoder and decoder. The biggest difference is the definition of query. 
In DETR, the decoder's queries represents $N$ learnable objects because DETR is designed for object detection.
However, in Trans2Seg, the queries represents $N$ learnable class prototypes, where each query represents one category. 
We could see that the minor change on query design could generalize Transformer architecture to apply to diverse vision tasks, such as object detection and semantic segmentation.

\section{Experiments}
\subsection{Implementation Details.}
We implement Trans2Seg with Pytorch. The ResNet-50~\cite{resnet} with dilation convolution at last stage. is adoped as the CNN extractor. 
For loss optimization, we use Adam optimizer with epsilon 1e-8 and weight decay 1e-4. Batch size is 8 per GPU. We set learning rate 1e-4 and decayed by the poly strategy~\cite{bisenet} for 50 epochs. We use 8 V100 GPUs for all experiments. 
For all CNN based methods, we random scale and crop the image to $480\times 480$ in training, and resize image to $513\times 513$ in inference, following common setting on PASCAL VOC~\cite{voc}.
For our Trans2Seg, we adopt transformer architecture and need to keep the shape of learned position embedding same in training/inference, so we directly resize the image to $512\times 512$.
Code has been released for community to follow.

\subsection{Ablation Studies.}
We use the FCN~\cite{fcn} as our baseline. FCN is a fully convolutional network with very simple design, and it is also a very classic semantic segmentation method.
First, we demonstrate that transformer encoder can build long range attention between pixels, which has much larger receptive field than CNN filters.
Second, we remove the CNN decoder in FCN and replace by our Transformer decoder, we design a set of learnable class prototypes as queries and show that this design further helps improve the accuracy.
Third, we verify our method with transformer at different scales.

\begin{table}[ht]
\small
  \begin{minipage}{1\linewidth}
    \centerline{\scalebox{1}{\input{tables/ab1}}}
    \caption{\textbf{Effectiveness of Transformer encoder and decoder.} `Trans.' indicates Transformer. `Enc.' and `Dec.' means encoder and decoder.}
      \label{tab:ab1}
    \end{minipage}
    \begin{minipage}{1\linewidth}
    \centerline{\scalebox{1}{\input{tables/ab3}}}
      \caption{\textbf{Performance of Transformer at different scales.} `e\{a\}-n\{b\}-m\{c\}' means the transformer with number of `a' embedding dims, `b' layers and `c' mlp ratio.}
      \label{tab:ab3}
    \end{minipage}
\end{table}

\textbf{Self-Attention of Transformer Encoder.} 
As shown in Figure~\ref{tab:ab1}, the FCN baseline without transformer encoder achieves 62.7\% mIoU, when adding transformer encoder, the mIoU directly improves 6.1\%, achieving 66.8\% mIoU. It demonstrates that the self-attention module in transformer encoder provides global receptive filed, which is better than CNN's local receptive field in transparent object segmentation.

\textbf{Category Prototypes of Transformer Decoder.}
In Figure~\ref{tab:ab1}, we verify the effectiveness of learnable category prototypes in transformer decoder.
In column 2, with traditional CNN decoder, the mIoU is 68.8\%. However, with our transformer decoder, the mIoU boosts up to 72.1\% with 3.3\% improvement. The strong performance benefits from the flexible representation that learnable category prototypes as queries to find corresponding pixels in feature map.

\textbf{Scale of Transformer.}
The scale of transformer is mainly influenced by three hyper-parameters: (1) embedding dim of feature. (2) number of attention layers. (3) mlp ratio in feed forward layer. We are interested in whether enlarge the model size can continuously improve performance. So we set three combinations, as shown in Figure~\ref{tab:ab3}. We can find that with the size of transformer increase, the mIoU first increase then decrease. We argue that if without massive data to pretrain, \textit{e.g.} BERT~\cite{bert} used large-scale nlp data, the transformer size is not the larger the better for our task.

\begin{figure}[pt]
    \centering
    \scalebox{0.13}{\includegraphics{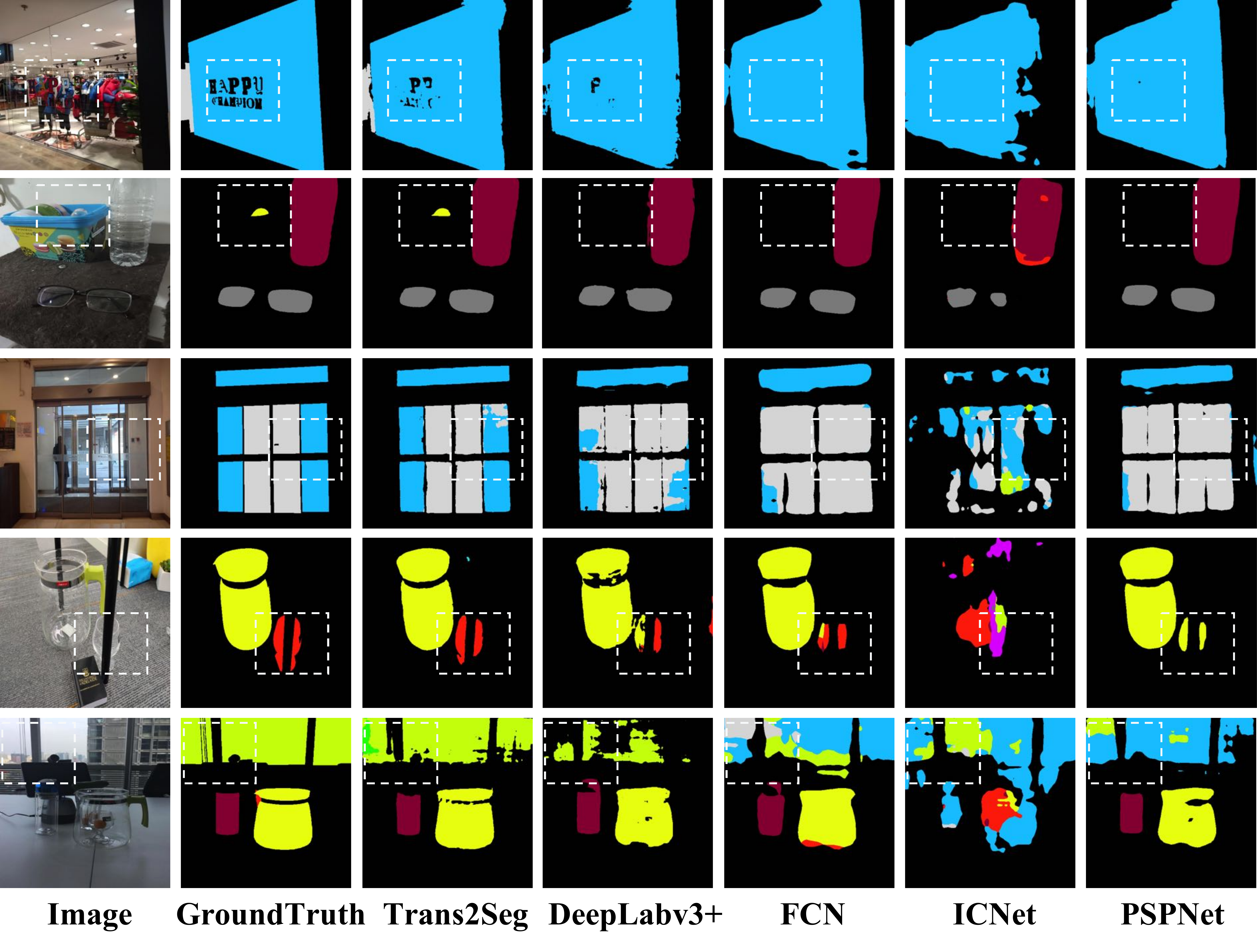}}
    \caption{\textbf{Visual comparison} of Trans2Seg to other CNN-based semantic segmentation methods. Our Trans2Seg clearly outperforms others thanks to the transformer's global receptive field and attention mechanism, especially in dash region. Zoom in for best view. Refer to supplementary materials for more visualized results.}
    \label{fig:vis}
\end{figure}

\subsection{Comparison to the state-of-the-art.}

\begin{table*}[h]
    \small
    \centering
    \scalebox{0.85}{\input{tables/experiment}}
    \vspace{10pt}
    \caption{\textbf{Evaluated state-of-the-art semantic segmentation methods.} Sorted by FLOPs. Our proposes Trans2Seg surpasses all the other methods in pixel accuracy and mean IoU, as well as most of the category IoUs (8 in 11).}
    \label{tab:sota}
    \vspace{-10pt}
\end{table*}

\begin{table}[t]
    \centering
    \setlength{\tabcolsep}{0.7mm}
    \input{tables/ade.tex}
    \caption{\textbf{Performance of Trans2Seg on ADE20K dataset.} Trans2Seg also works well on general semantic segmentation tasks. ``d8'' and ``d16'' means dilation 8 and 16, respectively. ``R50'' means ResNet-50 backbone.}
    \label{tab:ade}
\end{table}

We select more than 20 semantic segmentation methods~\cite{translab,deeplabv3+,dabnet,pspnet,ocnet,denseaspp,fcn,unet,bisenet,refinenet,hardnet,hrnet,fastscnn,contextnet,lednet,dunet,icnet,dabnet,fpenet,dfanet,danet,espnetv2} to evaluate on our Trans10K-v2 dataset, the methods selection largely follows the benchmark of TransLab~\cite{translab}. For fair comparsion, we train all the methods with 50 epochs.

Table~\ref{tab:sota} reports the overall quantitative comparison results on test set.
Our Trans2Seg achieves state-of-the-art 72.15\% mIoU and 94.14\% pixel ACC, significant outperforms other pure CNN-based methods. For example, our method is 2.1\% higher than TransLab, which is the previous SOTA method. We also find that our method tend to performs much better on small objects, such as `bottle' and 'eyeglass'~(10.0\% and 5.0\% higher than previous SOTA). We consider that the transformer's long range attention benefits the small transparent object segmentation.

In Figure~\ref{fig:vis}, we visualize the mask prediction of Trans2Seg and other CNN-based methods. We can find that benefit from transformer's large receptive field and attention mechanism, our method can distinguish background and different categories transparent objects much better than other methods, especially when multiple objects with different categories occurs in one image.
Moreover, our method can obtain high quality detail information,\textit{e.g.} boundary of object, and tiny transparent objects, while other CNN-based methods fail to do so. More results are shown in supplementary material.

\subsection{General Semantic Segmentation}
We try to transfer Trans2Seg on general semantic segmentation and it also achieves satisfied performance.

\textbf{Experiment Settings.}
We choose ADE20K~\cite{zhou2017scene}, a challenging scene parsing benchmark for semantic segmentation. ADE20K contains 150 fine-grained semantic categories, where there are 20210, 2000, and 3352 images for training, validation and, testing, respectively. We set learning rate to 2.5e-5 for ADE20K experiments. We train all models with 40k iterations with 8 images/GPU and 8 GPUs, and use single-scale test in inference. The data augmentation is same as DeeplabV3+~\cite{deeplabv3+}. 

\textbf{Results.}
As shown in Table~\ref{tab:ade}, compared with Semantic FPN and DeeplabV3+, our Trans2Seg achieves 39.7 mIoU, which is a satisfied performance. 
Our Trans2Seg verifies robust transfer ability on challenging general segmentation dataset.
Please note that we do not carefully tuned the hyper-parameters of Trans2Seg on ADE20K dataset. 
We are highly interested to design a better transformer-based general semantic segmentation pipeline in the future.

\section{Conclusion}
In this paper, we present a new fine-grained transparent object segmentation dataset with 11 common categories, termed Trans10K-v2, where the data is based on the previous Trans10K. We also discuss the challenging and practical of the proposed dataset. 
Moreover, we propose a transformer-based pipeline, termed Trans2Seg, to solve this challenging task. In Trans2Seg, the transformer encoder provides global receptive field, which is essential for transparent objects segmentation. In the transformer decoder, we model the segmentation as dictionary look up with a set of learnable queries, where each query represents one category. 
Finally, we evaluate more than 20 mainstream semantic segmentation methods and shows our Trans2Seg clearly surpass these CNN-based segmentation methods.

In the future, we are interested in exploring our Transformer encoder-decoder design on general segmentation tasks, such as Cityscapes and PASCAL VOC. We will also put more effort to solve transparent object segmentation task.

\clearpage

\section{Appendix}
\subsection{Detailed Dataset Information}
\subsubsection{More Visualized Demonstration of Trans10K-v2.}
\begin{figure}[ht]
     \includegraphics[width=0.95\linewidth]{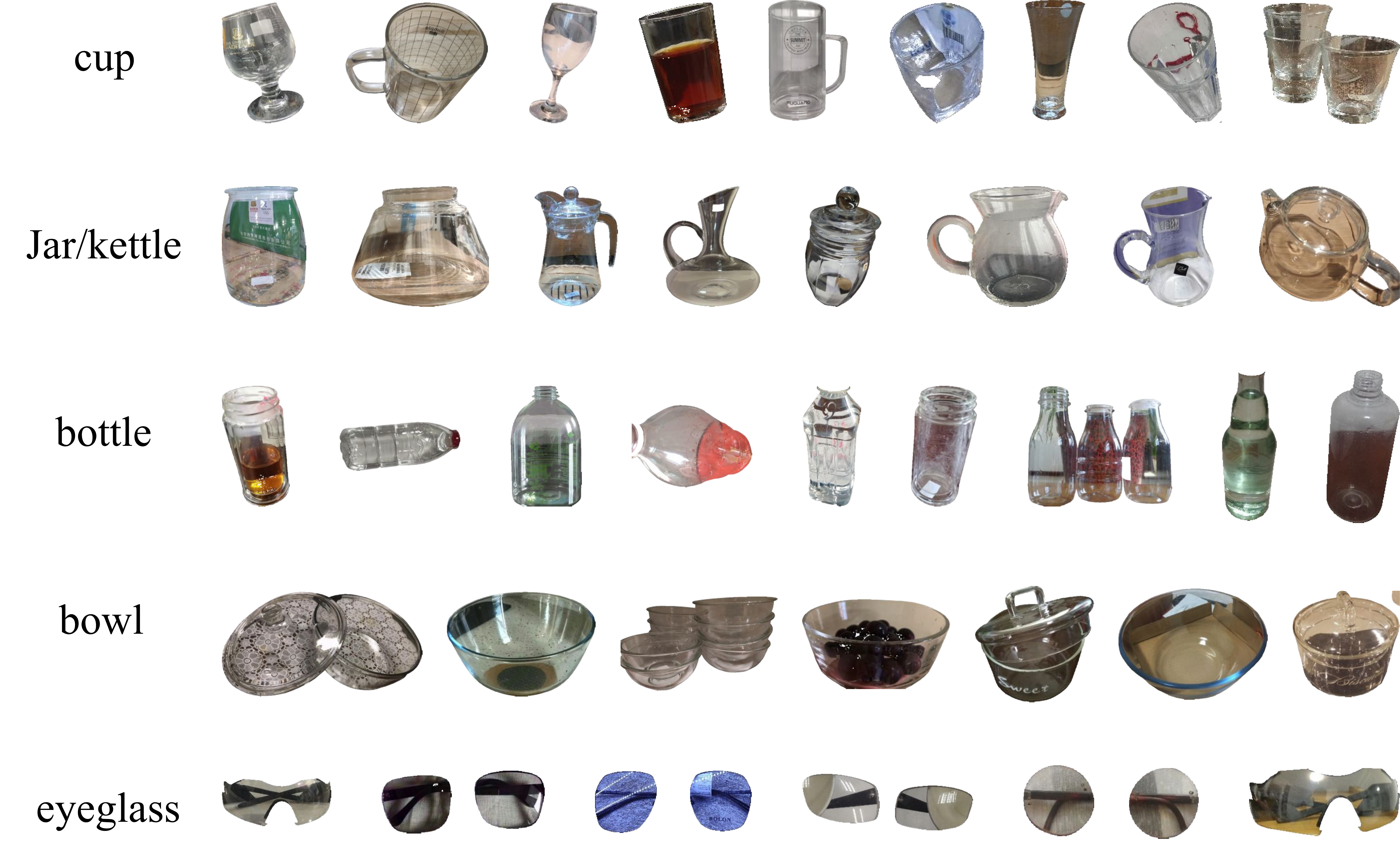}
     \caption{Cropped objects of 5 kinds of transparent things: cup, jar, bottle, bowl, eyeglass. Zoom in for the best view.}
     \label{fig:experiment_supp_thing}
\end{figure}

\begin{figure}[ht]
     \includegraphics[width=0.95\linewidth]{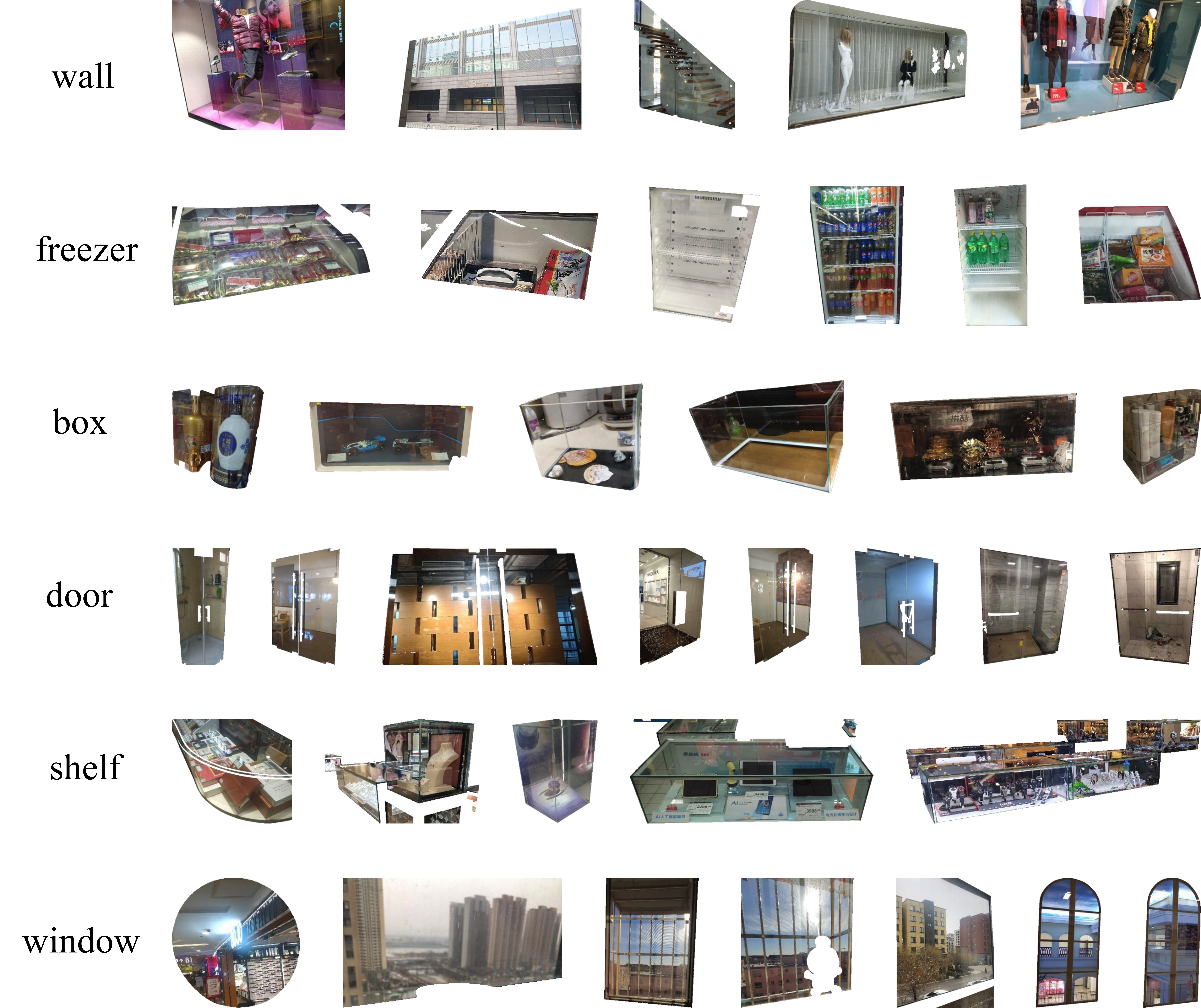}
     \caption{Cropped objects of 6 kinds of transparent stuff: wall, freezer, box, door, shelf, window. Zoom in for the best view.}
     \label{fig:experiment_supp_stuff}
\end{figure}

\begin{figure*}[t]
    \centering
    \scalebox{0.18}{\includegraphics{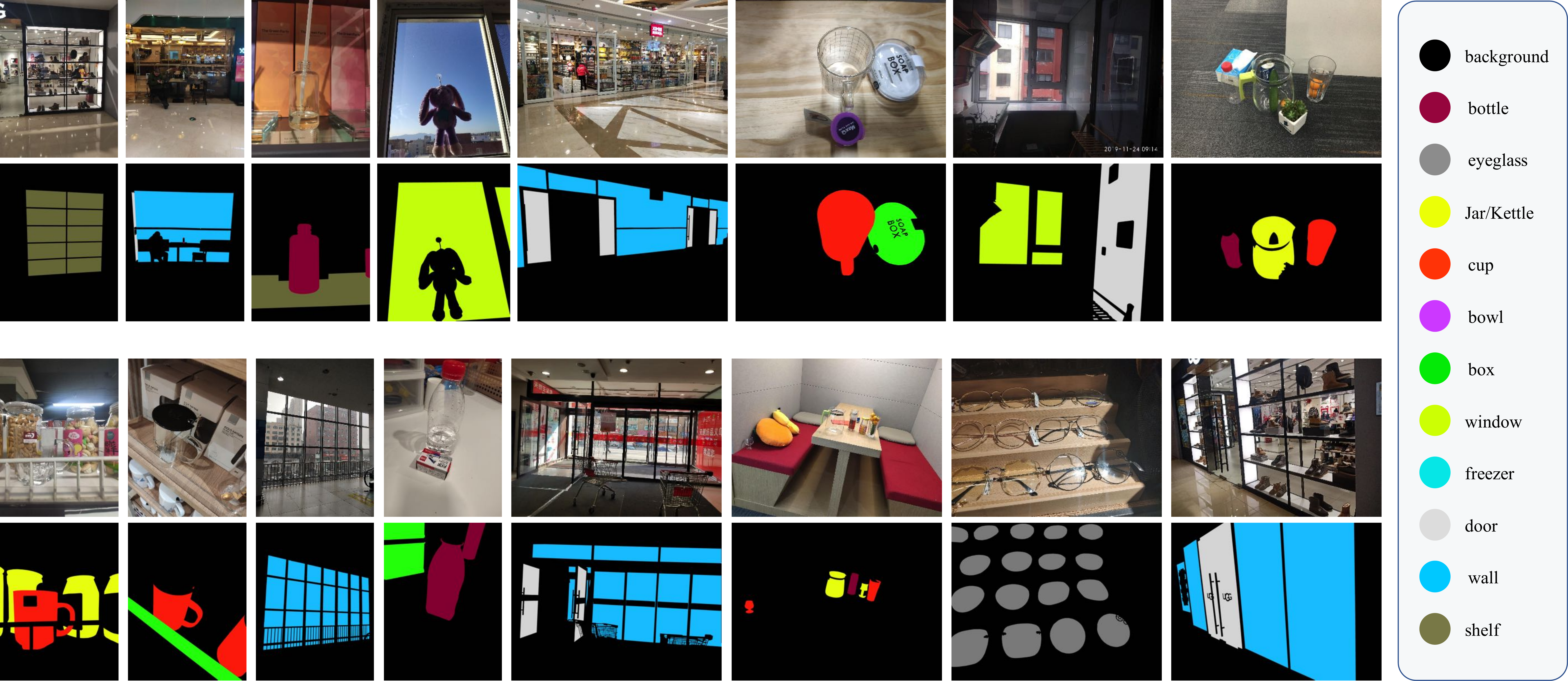}}
    \caption{\textbf{More images and corresponding high-quality masks in Trans10K-v2.} Our dataset is high diversity in scale, categories, pose, contrast, occlusion, and transparency. Zoom in for the best view.}
    \label{fig:dataset_supp}
\end{figure*}

In this section we show more visualized demonstrations to show the diversity and quality of Trans10K-v2. In
Figure~\ref{fig:experiment_supp_thing} and Figure~\ref{fig:experiment_supp_stuff}, we show more cropped objects to illustrate the high-diversity of the objects. We also show more images and ground-truth masks in Figure~\ref{fig:dataset_supp}. All images and transparent objects in Trans10K-v2 are selected from complex real-world scenarios that have large variations such as scale, viewpoint, contrast, occlusion, categories and transparency. From Figure~\ref{fig:dataset_supp}, we can also find that it is challenging for current semantic segmentation methods.

\subsubsection{Scene information}
We also provide each image with a scene label that represents where the objects located in. 
As shown in the upper part of Table~\ref{tab:scene_stat}, we list the statistics of the distribution in different scenarios of each category in detail. The distribution highly follows the distribution of our residential environments. For example, the cups, bowls, and bottles are mostly placed \textbf{on the desk}, while glass walls are often located in \textbf{mega-malls} or \textbf{office buildings}. 

\begin{figure*}[ht]
    \centering
    \scalebox{0.4}{\includegraphics{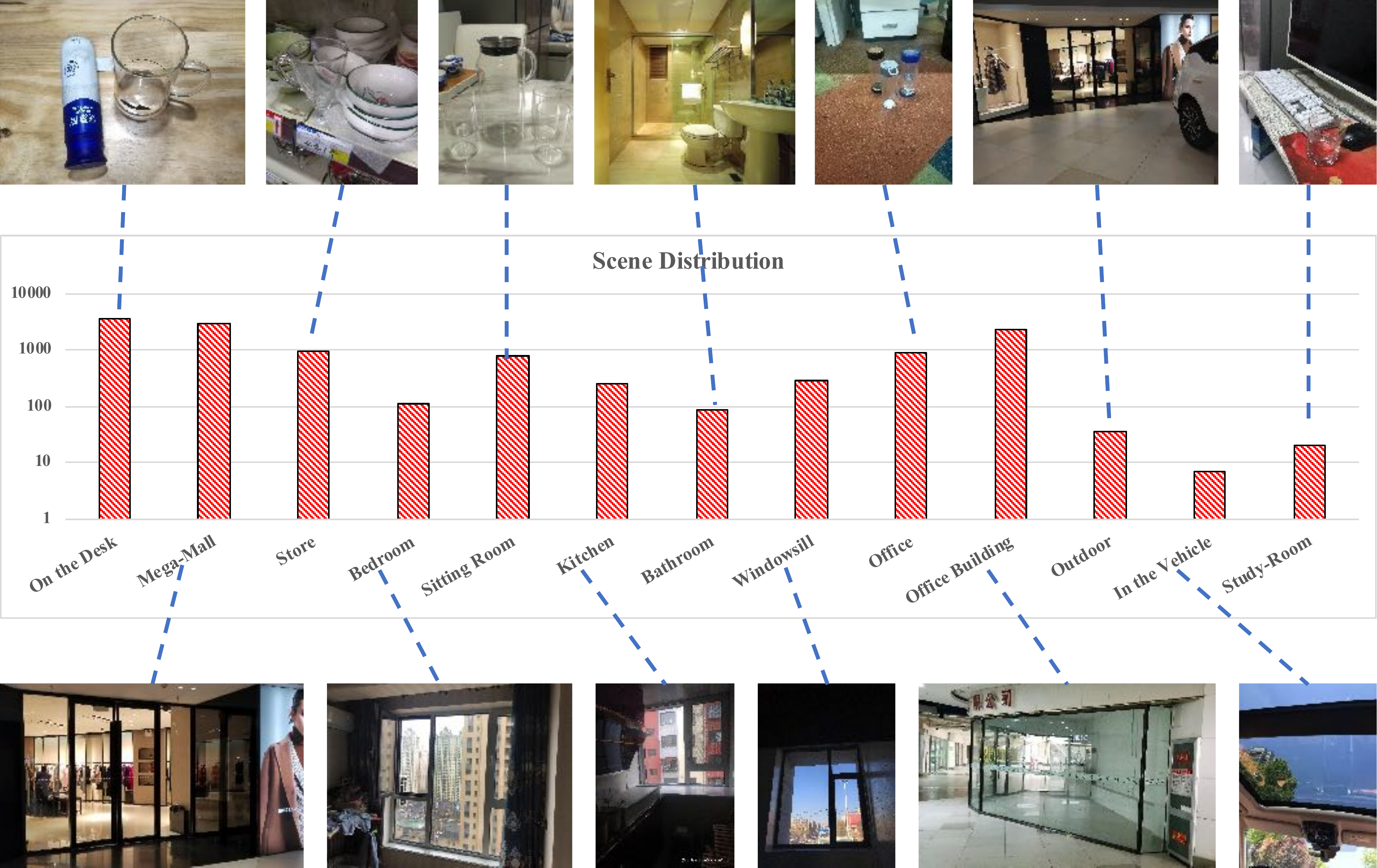}}
    \caption{The image number distribution and selected images of different scenes in Trans10K-v2. For better demonstration, the image number in vertical axis is listed as logarithmic.}
    \label{fig:scene_vis}
\end{figure*}

The visualized demonstration of our diverse scene distribution is shown in Figure~\ref{fig:scene_vis}. Trans10k-v2 contains abundant scenarios and we induce them into 13 categories: on the desk, mega-mall, store, bedroom, sitting room, kitchen, bathroom, windowsill, office, office building, outdoor, in the vehicle, study-room. This information is mainly used to demonstrate our abundant image distribution which could cover most of the common real-life scenarios. Each image is provided with a scene label.

\subsubsection{How Robots Deal with Transparent Objects}
Transparent objects are widespread in human residential environments, so the human-aiding robots find ways to deal with transparent objects. Some former robotic research illustrates the substantial value of solving this problem, mainly from grasping and navigation. This research primarily focuses on modifying the algorithm to deal with optical signals reflected from the transparent objects.

For the manipulator grasping, previous work mainly focuses on grabbing water cups.
\cite{detandrecon} propose an approach to reconstruct an approximate surface of the transparent cups and bottles by the internal sensory contradiction from two ToF (time of flight) images captured from an SR4k camera. The robot arm could grasp and manipulate the objects.
\cite{reachandgrasp} set up a BCI-robot platform to help patients suffering from limb muscle paralysis by grasping a glass cup for the patients. 
Starting from the point that the usual glass material absorbs light in specific wavelengths,
\cite{iros2018} propose the Depth Likelihood Volume (DLV), which uses a Monte Carlo object localization algorithm to help the Michigan Progress Fetch robot localize and manipulate translucent objects.

For the mobile robot navigation, some work also finds ways to exclude the side-effect of transparent stuff in residential scenarios.
\cite{icra2013} modify the standard occupancy grid algorithm during the procedure of autonomous-mapping robot localize transparent objects from certain angles.
\cite{laserfinder} design a novel scan matching algorithm by comparing all candidate distances scanned by the laser range finder penetrate and reflected from the glass walls.
\cite{multidata} use information fusion by combining a laser scanner and a sonar on an autonomous-mapping mobile robot to reduce the uncertainty caused by glass. 

We analyze how robots deal with transparent objects from previous work and grade them into 4 patterns:
\textbf{navigation, grasping, manipulation, human-aiding.}
Navigation and grasping are the two fundamental interactions between robots and objects. Manipulation happens on complex objects like windows, doors, or bottles with lids.
Human-aiding is the highest level of robot mission, and this kind of interaction always involve human, especially disabled patients.
From these 4 patterns, we can then analyze and categorize the transparent objects in respect to functions.

\subsubsection{Categorization Principle}
The 11 fine-grained categories are based on how the robots need to deal with transparent objects like avoiding or grasping or manipulating. 
For example, the goblet and cup are both open-mouthed and mainly used to drink water. These objects need to be grasped carefully since they do not have lids.  They have different interactive actions with the robots. So they are both categorized as cup.
We show the detailed demonstration of each category:
(1)~Shelf. Containing bookshelf, showcase, cabinet, etc. They mostly have sliding glass doors and are used to store goods. 
(2)~Freezer. Containing vending machine,
horizontal freezer, etc. They are electrical equipment and are used to storing drinks and food.
(3)~Door. Containing automatic glass door, standard glass door, etc. The doors are located in mega-mall, bathroom or office building. They are highly transparent and extensive. They could be used in navigation and helping disabled people pass through.
(4)~Wall. Glass walls look like doors. However, walls can not be opened. This clue should be perceived during mobile robots' mapping procedure. Glass walls are common in mega-mall and office buildings.
(5)~Window. Windows could be opened like glass doors but should not be traveled through. 
(6)~Box. Large boxes may not need to be grasped, but the manipulator robot needs to open the box and search for specific items.
(7)~Cup. We category all open-mouthed cups like goblets and regular cups into this category. Cups are used for drinking water. The manipulators need to grasp a cup carefully and be able to assist disabled people to drink water.
(8)~Bottle. Bottles are also used to drink water. But bottles have lids, so they need careful manipulation.
(9)~Eyeglass. Eyeglasses need careful grasping and manipulation to help disable people wear the eyeglasses.
(10)~Jar. This category contains jars, kettles and other transparent containers used to hold water, flavoring and food.
(11)~Bowl. Bowls are usually used to contain water or food. Different from jars, they do not have lids and need careful grasping.
The sample objects of these categories could be find in Figure~\ref{fig:dataset_supp}. We show the most common type of different categories by cropping the objects through masks.

As shown in the lower part of Table~\ref{tab:scene_stat}, we analyze and list the interactive patterns of all the 11 fine-grained categories of objects. Navigation is the basic interactive pattern of stuff and grasping is the basic interactive pattern of things. All the objects with some complex interactions need to be manipulated like the robots helping people open the shelf or window. Human-aiding is the highest level of interaction and it always involves patients. The patients need robots to help with opening the door, or feeding water by a cup or bottle.

\begin{figure*}[t]
    
    \centering
    \begin{subfigure}{0.4\textwidth}
    \includegraphics[width=1\linewidth]{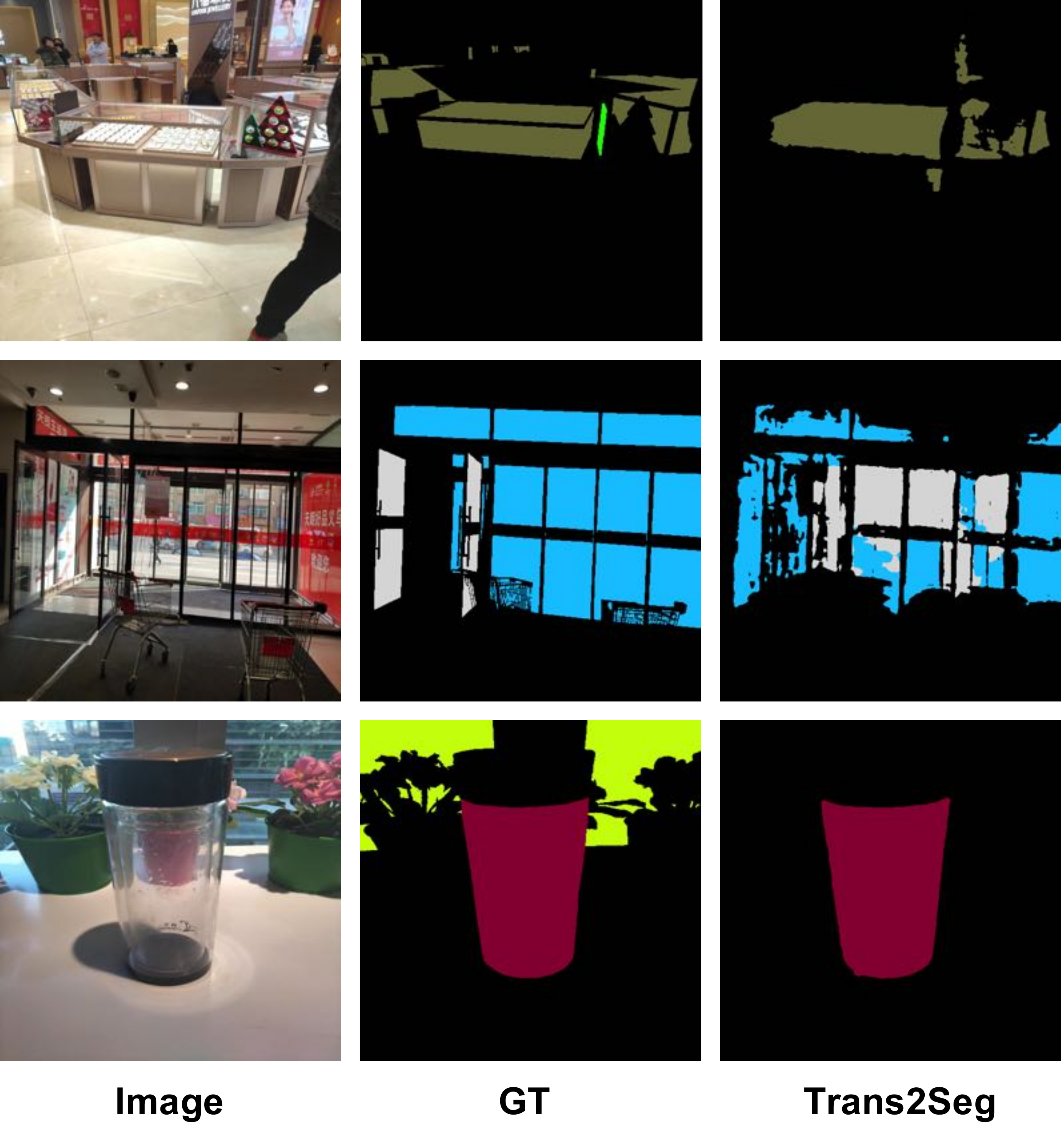}
    \small
    \caption{Occlusion and Crowd.}
    \label{fig:fail1}
    \end{subfigure}
    \hspace{30pt}
    \begin{subfigure}{0.4\textwidth}
     \includegraphics[width=1\linewidth]{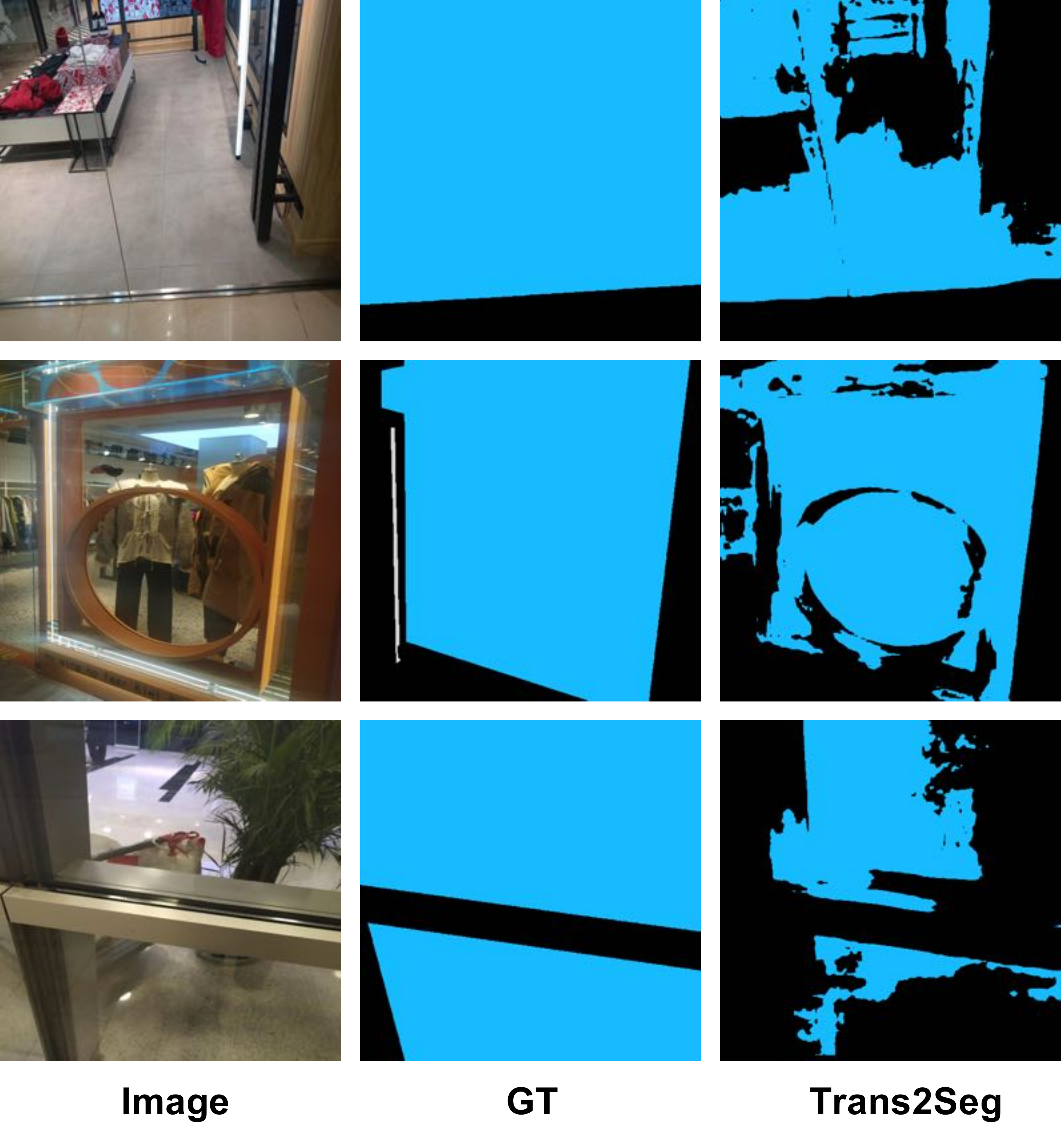}
     \small
     \caption{Extreme Transparency.}
    \label{fig:fail2}
    \end{subfigure}
    \small
    \caption{\textbf{Failure cases analysis.} Our Trans2Seg fails to segment transparent objects in some complex scenarios.}
    \label{fig:fail}
\end{figure*}

\subsection{More Visual Results Comparison.}
\begin{figure*}[ht]
    \centering
    \scalebox{0.19}{\includegraphics{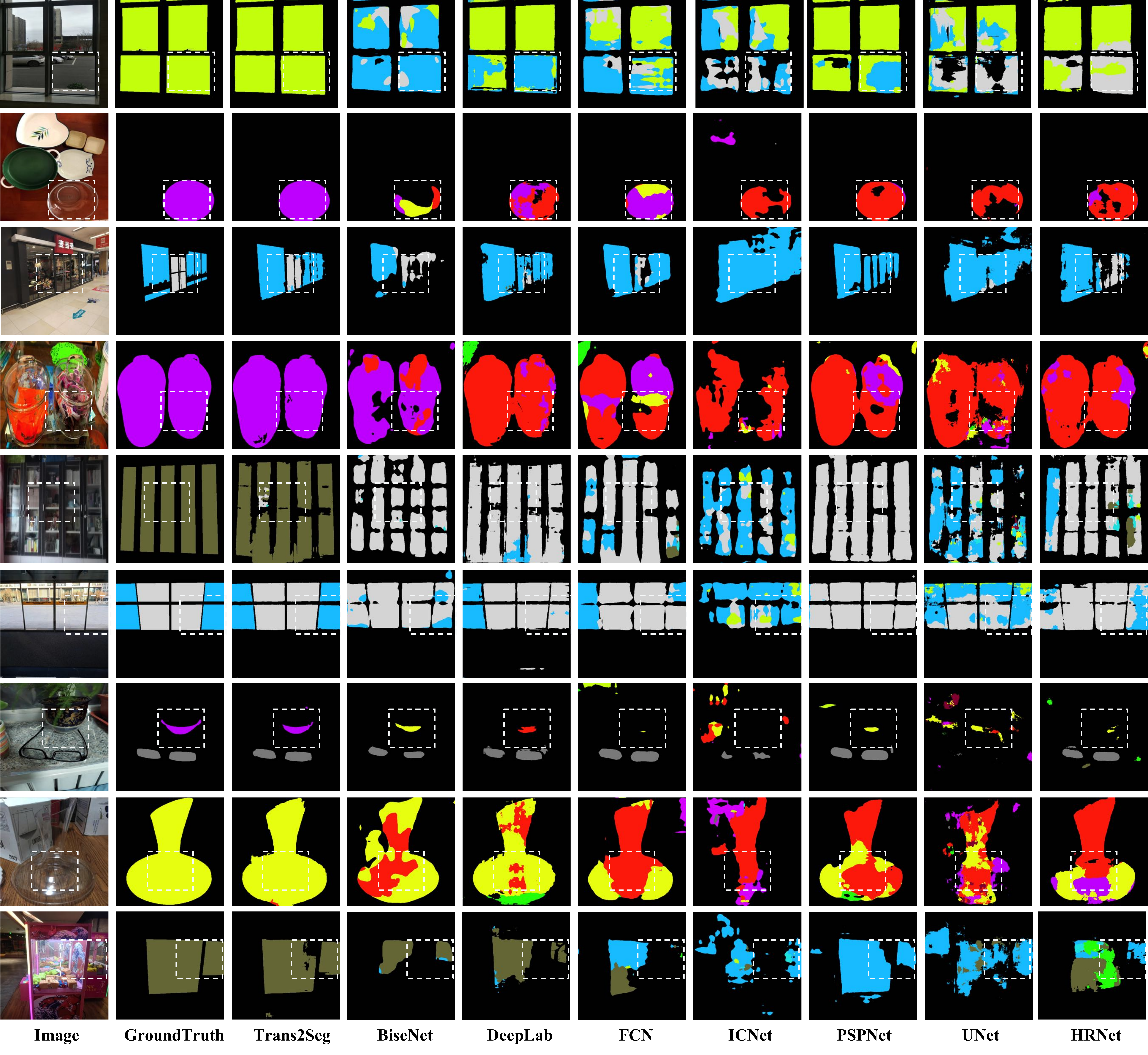}}
    \caption{\textbf{Visualized results of comparison with state-of-the-art methods.} Our Trans2Seg has the best mask prediction among all methods. Zoom in for the best view.}
    \label{fig:vis_supp}
\end{figure*}

In this section, we visualize more test examples produced by our Trans2Seg and other CNN-based methods on Trans10K-v2 dataset in Figure~\ref{fig:vis_supp}. 
From these results, we can easily observe that our Trans2Seg outputs very high-quality transparent object segmentation masks than other methods. 
Such strong results mainly benefit from the successfully introducing Transformer into transparent object segmentation, which is the lack in other CNN-based methods.

\begin{table*}[ht]
\centering
    \input{tables/scene_stat}
    \caption{The upper part of this table: the number of the scene. The lower part of this table: the interaction pattern of each category.}
\label{tab:scene_stat}
\end{table*}

\subsection{Failure Case Analysis}
As shown in Figure~\ref{fig:fail}, our method also has some limitations. For instance, in Figure~\ref{fig:fail}~(a), when transparent objects are occluded by different categories, our method would confuse and fail to segment part of the items. 
In Figure~\ref{fig:fail}~(b), when the objects are of extreme transparency, our method would also confuse and output wrong segmentation results. In such a case, even humans would also fail to distinguish these transparent objects.

\clearpage

\bibliographystyle{named}
\bibliography{ijcai21}
\end{document}

%% file: tables/stat2.tex
\begin{tabular}{p{60pt}<{\centering}p{17pt}<{\centering}p{17pt}<{\centering}p{17pt}<{\centering}p{17pt}<{\centering}p{17pt}<{\centering}p{30pt}<{\centering}p{17pt}<{\centering}p{17pt}<{\centering}p{17pt}<{\centering}p{17pt}<{\centering}p{22pt}<{\centering}}
\toprule

Trans10Kv2 &  shelf & door & wall & box &  freezer & window & cup & bottle  & jar&  bowl & eyeglass\\ \midrule
image num & 280 & 1572 & 3059 & 603 & 90 & 501 &  3315 & 1472 & 997 & 340 & 410 \\
CMCC & 3.36  & 5.19 & 5.61 & 2.57 & 3.36 & 4.27 &  1.97 & 1.82 & 1.99 & 1.31& 2.56  \\
pixel ratio(\%) & 2.49  & 9.23  & 38.42  & 3.67   & 1.02 & 4.28  &  22.61  & 6.23   & 6.75  & 3.67& 0.78   \\
\bottomrule
\end{tabular}

%% file: tables/ab1.tex
\begin{tabular}{p{20pt}<{\centering}p{45pt}<{\centering}p{45pt}<{\centering}p{40pt}<{\centering}p{25pt}<{\centering}}
\toprule
id & Trans. Enc. & Trans. Dec. & CNN Dec. & mIoU \\ 
\midrule
0  & $\times$           & $\times$           & \checkmark        & 62.7 \\ 
1  & \checkmark           & $\times$           & \checkmark        & 68.8 \\ 
2  & \checkmark           & \checkmark           & $\times$        & 72.1 \\ \bottomrule
\end{tabular}

%% file: tables/ab3.tex
\begin{tabular}{p{30pt}<{\raggedright}p{60pt}<{\centering}p{30pt}<{\centering}p{30pt}<{\centering}p{30pt}<{\centering}}
\toprule
Scale  & hyper-param. & GFlops    & MParams   & mIoU     \\ \midrule
small  & e128-n1-m2   & 40.9      & 30.5  & 69.2     \\
medium & e256-n4-m3   & 49.0      & 56.2      & 72.1     \\
large  & e768-n12-m4  & 221.8     & 327.5     & 70.3 \\\bottomrule
\end{tabular}

%% file: tables/experiment.tex
\begin{tabular}{p{60pt}<{\raggedright}p{21pt}<{\centering}p{21pt}<{\centering}p{27pt}<{\centering}p{21pt}<{\centering}p{21pt}<{\centering}p{21pt}<{\centering}p{21pt}<{\centering}p{21pt}<{\centering}p{21pt}<{\centering}p{21pt}<{\centering}p{21pt}<{\centering}p{21pt}<{\centering}p{21pt}<{\centering}p{21pt}<{\centering}p{21pt}<{\centering}}
\toprule
\multirow{2}{*}{\large Method} & \multirow{2}{*}{FLOPs} & \multirow{2}{*}{ACC~$\uparrow$} & \multirow{2}{*}{mIoU~$\uparrow$} & \multicolumn{12}{c}{\rule{0pt}{10pt}Category IoU~$\uparrow$} \\ \cline{5-16} 

 \rule{0pt}{10pt} &  \multicolumn{1}{|c|}{} &  &  \multicolumn{1}{c|}{} & bg & shelf  & Jar  & freezer & window & door & eyeglass &cup & wall & bowl & bottle &box  \\ \midrule
 
 \rule{0pt}{10pt} FPENet  &\multicolumn{1}{|c|}{0.76} & 70.31 & \multicolumn{1}{c|}{10.14} & 74.97 & 0.01 & 0.00 & 0.02 & 2.11 & 2.83 & 0.00 & 16.84 & 24.81 & 0.00 & 0.04 & 0.00 \\
 
\rule{0pt}{10pt} ESPNetv2  & \multicolumn{1}{|c|}{0.83} & 73.03 & \multicolumn{1}{c|}{12.27} & 78.98 & 0.00 & 0.00 & 0.00 & 0.00 & 6.17 & 0.00 & 30.65 & 37.03 & 0.00 & 0.00 & 0.00 \\

\rule{0pt}{10pt} ContextNet & \multicolumn{1}{|c|}{0.87} & 86.75 & \multicolumn{1}{c|}{46.69}   & 89.86 & 23.22 & 34.88 & 32.34 & 44.24 & 42.25 & 50.36 &  65.23 & 60.00 & 43.88 & 53.81 & 20.17 \\

\rule{0pt}{10pt} FastSCNN  & \multicolumn{1}{|c|}{1.01} & 88.05 & \multicolumn{1}{c|}{51.93} & 90.64 & 32.76 & 41.12 & 47.28 &  47.47 & 44.64 & 48.99 & 67.88 & 63.80 & 55.08 & 58.86 & 24.65 \\

\rule{0pt}{10pt} DFANet  & \multicolumn{1}{|c|}{1.02} & 85.15 & \multicolumn{1}{c|}{42.54} & 88.49 & 26.65 & 27.84 & 28.94 & 46.27 & 39.47 & 33.06 & 58.87 & 59.45 & 43.22 & 44.87 & 13.37 \\
\rule{0pt}{10pt} ENet  & \multicolumn{1}{|c|}{2.09} & 71.67 & \multicolumn{1}{c|}{8.50} & 79.74 & 0.00 &  0.00 & 0.00 & 0.00 & 0.00 & 0.00 & 0.00 & 22.25 & 0.00 & 0.00 & 0.00 \\


\rule{0pt}{10pt} HRNet\_w18  & \multicolumn{1}{|c|}{4.20} & 89.58 & \multicolumn{1}{c|}{54.25} & 92.47 & 27.66 & 45.08 & 40.53 & 45.66 & 45.00 & 68.05 & 73.24 & 64.86& 52.85 & 62.52 & 33.02   \\
\rule{0pt}{10pt} HardNet  & \multicolumn{1}{|c|}{4.42} & 90.19 & \multicolumn{1}{c|}{56.19} & 92.87 & 34.62 & 47.50 & 42.40 & 49.78 & 49.19 & 62.33 & 72.93 & 68.32 & 58.14 & 65.33 & 30.90 \\
\rule{0pt}{10pt} DABNet  & \multicolumn{1}{|c|}{5.18} & 77.43 & \multicolumn{1}{c|}{15.27} & 81.19 & 0.00 & 0.09 & 0.00 & 4.10 & 10.49 & 0.00 & 36.18 & 42.83& 0.00 & 8.30 & 0.00  \\
\rule{0pt}{10pt} LEDNet  & \multicolumn{1}{|c|}{6.23} & 86.07 & \multicolumn{1}{c|}{46.40} & 88.59 & 28.13 & 36.72 & 32.45 & 43.77 & 38.55 & 41.51 & 64.19 & 60.05 & 42.40 & 53.12 & 27.29 \\
\rule{0pt}{10pt} ICNet  & \multicolumn{1}{|c|}{10.64} & 78.23 &\multicolumn{1}{c|}{23.39} & 83.29 & 2.96 & 4.91 & 9.33 & 19.24 & 15.35 & 24.11 & 44.54 & 41.49 & 7.58 & 27.47 & 3.80  \\
\rule{0pt}{10pt} BiSeNet  & \multicolumn{1}{|c|}{19.91} & 89.13 & \multicolumn{1}{c|}{58.40} & 90.12 & 39.54 &53.71  &50.90  & 46.95& 44.68 & 64.32 & 72.86 &63.57 &61.38 & 67.88 &44.85  \\

\midrule

\rule{0pt}{10pt} DenseASPP  & \multicolumn{1}{|c|}{36.20} & 90.86 & \multicolumn{1}{c|}{63.01} & 91.39 & 42.41 & 60.93 & 64.75 & 48.97 & 51.40 & 65.72 & 75.64 & 67.93 & 67.03 & 70.26 & 49.64  \\
\rule{0pt}{10pt} DeepLabv3+ & \multicolumn{1}{|c|}{37.98} & 92.75 & \multicolumn{1}{c|}{68.87}  & 93.82 & 51.29 &64.65  & 65.71 & 55.26 & 57.19 & 77.06 & 81.89 & 72.64 &70.81  & 77.44 & \textbf{58.63}\\
\rule{0pt}{10pt} FCN  & \multicolumn{1}{|c|}{42.23} & 91.65 & \multicolumn{1}{c|}{62.75} & 93.62 & 38.84 & 56.05 & 58.76 & 46.91 & 50.74 & 82.56 &78.71 & 68.78 & 57.87 & 73.66 & 46.54 \\
\rule{0pt}{10pt} OCNet  & \multicolumn{1}{|c|}{43.31} & 92.03 & \multicolumn{1}{c|}{66.31} & 93.12 & 41.47 & 63.54 & 60.05 & 54.10 & 51.01 & 79.57 & 81.95 & 69.40 & 68.44 & 78.41 & 54.65\\
\rule{0pt}{10pt} RefineNet  & \multicolumn{1}{|c|}{44.56} & 87.99 & \multicolumn{1}{c|}{58.18} & 90.63 & 30.62 & 53.17 & 55.95 & 42.72 &46.59 & 70.85 & 76.01 & 62.91 & 57.05 & 70.34 & 41.32   \\

\rule{0pt}{10pt} Translab  & \multicolumn{1}{|c|}{61.31} & 92.67  & \multicolumn{1}{c|}{69.00} & 93.90 & \textbf{54.36} & 64.48 & 65.14 & 54.58 & 57.72 & 79.85 & 81.61 &72.82 & 69.63 & 77.50 & 56.43 \\ 

\rule{0pt}{10pt} DUNet  & \multicolumn{1}{|c|}{123.69} & 90.67 & \multicolumn{1}{c|}{59.01} & 93.07 & 34.20 & 50.95 & 54.96 & 43.19 & 45.05 & 79.80 & 76.07 & 65.29 & 54.33 & 68.57 & 42.64   \\
\rule{0pt}{10pt} UNet  & \multicolumn{1}{|c|}{124.55} & 81.90 & \multicolumn{1}{c|}{29.23} & 86.34 & 8.76  & 15.18  & 19.02 & 27.13 & 24.73 & 17.26 & 53.40 & 47.36 & 11.97 & 37.79 & 1.77 \\

\rule{0pt}{10pt} DANet  & \multicolumn{1}{|c|}{198.00} & 92.70  &\multicolumn{1}{c|}{68.81} & 93.69 & 47.69 & 66.05  & 70.18 & 53.01 & 56.15 & 77.73 & 82.89 & 72.24 & 72.18 & 77.87 & 56.06 \\

\rule{0pt}{10pt} PSPNet  & \multicolumn{1}{|c|}{187.03} & 92.47 & \multicolumn{1}{c|}{68.23} & 93.62  & 50.33 & 64.24 & \textbf{70.19} & 51.51 & 55.27 & 79.27 & 81.93 & 71.95 &   68.91 & 77.13 & 54.43 \\

\midrule 
\rule{0pt}{10pt} \textbf{Trans2Seg} &  \multicolumn{1}{|c|}{49.03} &  \textbf{94.14}  & \multicolumn{1}{c|}{\textbf{72.15}} & \textbf{95.35} & 53.43 & \textbf{67.82} & 64.20 & \textbf{59.64} & \textbf{60.56} & \textbf{88.52} & \textbf{86.67} & \textbf{75.99} & \textbf{73.98} & \textbf{82.43} & 57.17 \\ 
\bottomrule
\end{tabular}

%% file: tables/ade.tex
\begin{tabular}{l|c|c|c}
	Method & \#Param (M) &GFLOPs & mIoU (\%)   \\
	\hline
	R50-SemanticFPN & 28.5 & 45.6 & 36.7 \\
	R50-d8+DeeplabV3+  &26.8&120.5 &41.5\\
	R50-d16+DeeplabV3+  &26.8&45.5 &40.6\\
	\hline
	R50-d16+Trans2Seg  & 56.1&79.3 & 39.7\\

\end{tabular}

%% file: tables/scene_stat.tex
\begin{tabular}{p{90pt}<{\centering}p{22pt}<{\centering}p{22pt}<{\centering}p{22pt}<{\centering}p{22pt}<{\centering}p{22pt}<{\centering}p{22pt}<{\centering}p{9pt}<{\centering}p{22pt}<{\centering}p{22pt}<{\centering}p{22pt}<{\centering}p{22pt}<{\centering}p{22pt}<{\centering}}
\toprule
\multirow{2}{*}{$\frac{Scene/Category}{Interaction}$} & \multicolumn{6}{c}{Stuff}                          &  & \multicolumn{5}{c}{Things}                   \\\cline{2-7}\cline{9-13}
\rule{0pt}{9pt} &  shelf & freezer & door & wall & window & box &  & cup & bottle & eyeglass & jar&  bowl \\ \cline{1-7}\cline{9-13}
\rule{0pt}{9pt}on the desk                     & 3     & 0       & 0          & 2    & 4      & 227 &  & 1946 & 834    & 239      & 302         & 117 \\
\rule{0pt}{9pt}mega-mall                       & 219   & 35      & 450        & 1762 & 76     & 128 &  & 169  & 36     & 75       & 94          & 14  \\
\rule{0pt}{9pt}store                           & 13    & 36      & 5          & 19   & 3      & 75  &  & 444  & 111    & 1        & 175         & 57  \\
\rule{0pt}{9pt}bedroom                         & 6     & 0       & 4          & 9    & 23     & 2   &  & 23   & 33     & 6        & 6           & 1   \\
\rule{0pt}{9pt}living room                     & 10    & 0       & 7          & 14   & 19     & 52  &  & 310  & 167    & 25       & 139         & 67  \\
\rule{0pt}{9pt}kitchen                         & 0     & 8       & 6          & 4    & 4      & 19  &  & 79   & 23     & 0        & 46          & 66  \\
\rule{0pt}{9pt}bathroom                        & 0     & 0       & 33         & 31   & 8      & 4   &  & 5    & 3      & 4        & 0           & 2   \\
\rule{0pt}{9pt}windowsill                      & 0     & 0       & 0          & 31   & 209    & 4   &  & 17   & 8      & 8        & 17          & 2   \\
\rule{0pt}{9pt}office room                       & 15    & 7       & 25         & 43   & 12     & 84  &  & 298  & 235    & 51       & 158         & 2   \\
\rule{0pt}{9pt}office building                 & 8     & 3       & 1021       & 1107 & 131    & 5   &  & 1    & 5      & 0        & 2           & 0   \\
\rule{0pt}{9pt}outdoor                         & 0     & 0       & 13         & 20   & 2      & 0   &  & 0    & 2      & 0        & 0           & 0   \\
\rule{0pt}{9pt}in the vehicle                  & 0     & 0       & 2          & 0    & 1      & 0   &  & 4    & 0      & 0        & 0           & 0   \\
\rule{0pt}{9pt}study-room                      & 4     & 0       & 3          & 2    & 4      & 1   &  & 4    & 1      & 0        & 2           & 0  \\\midrule
\rule{0pt}{9pt} navigation &  \checkmark &  \checkmark & \checkmark  & \checkmark  & \checkmark & \checkmark  &  & & & & & \\
\rule{0pt}{9pt} grasping & & & & & & & &\checkmark  & \checkmark & \checkmark &\checkmark  & \checkmark \\
\rule{0pt}{9pt} manipulation & \checkmark  & \checkmark  &  \checkmark  &  & \checkmark  & \checkmark  & \ &  &  \checkmark  &\checkmark  & \checkmark   &   \\
\rule{0pt}{9pt} human-aiding & & &\checkmark  & & & &  & \checkmark &\checkmark  &\checkmark  & & \checkmark \\\bottomrule
\end{tabular}

%% file: arxiv.bbl
\begin{thebibliography}{}

\bibitem[\protect\citeauthoryear{Carion \bgroup \em et al.\egroup
  }{2020}]{DETR}
Nicolas Carion, Francisco Massa, Gabriel Synnaeve, Nicolas Usunier, Alexander
  Kirillov, and Sergey Zagoruyko.
\newblock {End-to-End} object detection with transformers.
\newblock In {\em ECCV}, 2020.

\bibitem[\protect\citeauthoryear{Chao \bgroup \em et al.\egroup
  }{2019}]{hardnet}
Ping Chao, Chao-Yang Kao, Yu-Shan Ruan, Chien-Hsiang Huang, and Youn-Long Lin.
\newblock Hardnet: A low memory traffic network.
\newblock In {\em ICCV}, 2019.

\bibitem[\protect\citeauthoryear{Chen \bgroup \em et al.\egroup }{2014}]{crf}
Liang-Chieh Chen, George Papandreou, Iasonas Kokkinos, Kevin Murphy, and Alan~L
  Yuille.
\newblock Semantic image segmentation with deep convolutional nets and fully
  connected crfs.
\newblock {\em arXiv}, 2014.

\bibitem[\protect\citeauthoryear{Chen \bgroup \em et al.\egroup
  }{2017}]{deeplab}
Liang-Chieh Chen, George Papandreou, Iasonas Kokkinos, Kevin Murphy, and Alan~L
  Yuille.
\newblock Deeplab: Semantic image segmentation with deep convolutional nets,
  atrous convolution, and fully connected crfs.
\newblock {\em TPAMI}, 2017.

\bibitem[\protect\citeauthoryear{Chen \bgroup \em et al.\egroup
  }{2018a}]{tomnet}
Guanying Chen, Kai Han, and Kwan{-}Yee~K. Wong.
\newblock Tom-net: Learning transparent object matting from a single image.
\newblock In {\em CVPR}, 2018.

\bibitem[\protect\citeauthoryear{Chen \bgroup \em et al.\egroup
  }{2018b}]{deeplab2}
Liang-Chieh Chen, Yukun Zhu, George Papandreou, Florian Schroff, and Hartwig
  Adam.
\newblock Encoder-decoder with atrous separable convolution for semantic image
  segmentation.
\newblock In {\em ECCV}, 2018.

\bibitem[\protect\citeauthoryear{Chen \bgroup \em et al.\egroup
  }{2018c}]{deeplabv3+}
Liang-Chieh Chen, Yukun Zhu, George Papandreou, Florian Schroff, and Hartwig
  Adam.
\newblock Encoder-decoder with atrous separable convolution for semantic image
  segmentation.
\newblock In {\em ECCV}, 2018.

\bibitem[\protect\citeauthoryear{Chen \bgroup \em et al.\egroup
  }{2020}]{chen2020pre}
Hanting Chen, Yunhe Wang, Tianyu Guo, Chang Xu, Yiping Deng, Zhenhua Liu, Siwei
  Ma, Chunjing Xu, Chao Xu, and Wen Gao.
\newblock Pre-trained image processing transformer.
\newblock {\em arXiv preprint arXiv:2012.00364}, 2020.

\bibitem[\protect\citeauthoryear{Devlin \bgroup \em et al.\egroup
  }{2019}]{bert}
Jacob Devlin, Ming{-}Wei Chang, Kenton Lee, and Kristina Toutanova.
\newblock {BERT:} pre-training of deep bidirectional transformers for language
  understanding.
\newblock In Jill Burstein, Christy Doran, and Thamar Solorio, editors, {\em
  Proceedings of the 2019 Conference of the North American Chapter of the
  Association for Computational Linguistics: Human Language Technologies,
  {NAACL-HLT} 2019, Minneapolis, MN, USA, June 2-7, 2019, Volume 1 (Long and
  Short Papers)}, pages 4171--4186. Association for Computational Linguistics,
  2019.

\bibitem[\protect\citeauthoryear{Dosovitskiy \bgroup \em et al.\egroup
  }{2020}]{dosovitskiy2020image}
Alexey Dosovitskiy, Lucas Beyer, Alexander Kolesnikov, Dirk Weissenborn,
  Xiaohua Zhai, Thomas Unterthiner, Mostafa Dehghani, Matthias Minderer, Georg
  Heigold, Sylvain Gelly, et~al.
\newblock An image is worth 16x16 words: Transformers for image recognition at
  scale.
\newblock {\em arXiv preprint arXiv:2010.11929}, 2020.

\bibitem[\protect\citeauthoryear{Everingham and Winn}{2011}]{voc}
Mark Everingham and John Winn.
\newblock The pascal visual object classes challenge 2012 (voc2012) development
  kit.
\newblock {\em Pattern Analysis, Statistical Modelling and Computational
  Learning, Tech. Rep}, 2011.

\bibitem[\protect\citeauthoryear{Foster \bgroup \em et al.\egroup
  }{2013}]{icra2013}
Paul Foster, Zhenghong Sun, Jong~Jin Park, and Benjamin Kuipers.
\newblock Visagge: Visible angle grid for glass environments.
\newblock In {\em 2013 IEEE International Conference on Robotics and
  Automation}, pages 2213--2220. IEEE, 2013.

\bibitem[\protect\citeauthoryear{Fu \bgroup \em et al.\egroup }{2019}]{danet}
Jun Fu, Jing Liu, Haijie Tian, Yong Li, Yongjun Bao, Zhiwei Fang, and Hanqing
  Lu.
\newblock Dual attention network for scene segmentation.
\newblock In {\em Proceedings of the IEEE Conference on Computer Vision and
  Pattern Recognition}, pages 3146--3154, 2019.

\bibitem[\protect\citeauthoryear{Gadde \bgroup \em et al.\egroup
  }{2016}]{gadde2016superpixel}
Raghudeep Gadde, Varun Jampani, Martin Kiefel, Daniel Kappler, and Peter~V
  Gehler.
\newblock Superpixel convolutional networks using bilateral inceptions.
\newblock In {\em ECCV}, 2016.

\bibitem[\protect\citeauthoryear{Gehring \bgroup \em et al.\egroup
  }{2017}]{posencoding}
Jonas Gehring, Michael Auli, David Grangier, Denis Yarats, and Yann~N Dauphin.
\newblock Convolutional sequence to sequence learning.
\newblock {\em arXiv preprint arXiv:1705.03122}, 2017.

\bibitem[\protect\citeauthoryear{Han \bgroup \em et al.\egroup
  }{2020}]{han2020survey}
Kai Han, Yunhe Wang, Hanting Chen, Xinghao Chen, Jianyuan Guo, Zhenhua Liu,
  Yehui Tang, An~Xiao, Chunjing Xu, Yixing Xu, et~al.
\newblock A survey on visual transformer.
\newblock {\em arXiv preprint arXiv:2012.12556}, 2020.

\bibitem[\protect\citeauthoryear{He \bgroup \em et al.\egroup }{2016}]{resnet}
Kaiming He, Xiangyu Zhang, Shaoqing Ren, and Jian Sun.
\newblock Deep residual learning for image recognition.
\newblock In {\em CVPR}, 2016.

\bibitem[\protect\citeauthoryear{Jin \bgroup \em et al.\egroup }{2019}]{dunet}
Qiangguo Jin, Zhaopeng Meng, Tuan~D Pham, Qi~Chen, Leyi Wei, and Ran Su.
\newblock Dunet: A deformable network for retinal vessel segmentation.
\newblock {\em Knowledge-Based Systems}, 2019.

\bibitem[\protect\citeauthoryear{Kim and Chung}{2016}]{laserfinder}
Jiwoong Kim and Woojin Chung.
\newblock Localization of a mobile robot using a laser range finder in a
  glass-walled environment.
\newblock {\em IEEE Transactions on Industrial Electronics}, 63(6):3616--3627,
  2016.

\bibitem[\protect\citeauthoryear{Klank \bgroup \em et al.\egroup
  }{2011}]{detandrecon}
Ulrich Klank, Daniel Carton, and Michael Beetz.
\newblock Transparent object detection and reconstruction on a mobile platform.
\newblock In {\em IEEE International Conference on Robotics \& Automation},
  2011.

\bibitem[\protect\citeauthoryear{Li \bgroup \em et al.\egroup }{2019a}]{dabnet}
Gen Li, Inyoung Yun, Jonghyun Kim, and Joongkyu Kim.
\newblock Dabnet: Depth-wise asymmetric bottleneck for real-time semantic
  segmentation.
\newblock {\em arXiv}, 2019.

\bibitem[\protect\citeauthoryear{Li \bgroup \em et al.\egroup }{2019b}]{dfanet}
Hanchao Li, Pengfei Xiong, Haoqiang Fan, and Jian Sun.
\newblock Dfanet: Deep feature aggregation for real-time semantic segmentation.
\newblock In {\em Proceedings of the IEEE Conference on Computer Vision and
  Pattern Recognition}, pages 9522--9531, 2019.

\bibitem[\protect\citeauthoryear{Lin \bgroup \em et al.\egroup
  }{2016}]{lin2016efficient}
Guosheng Lin, Chunhua Shen, Anton Van Den~Hengel, and Ian Reid.
\newblock Efficient piecewise training of deep structured models for semantic
  segmentation.
\newblock In {\em CVPR}, 2016.

\bibitem[\protect\citeauthoryear{Lin \bgroup \em et al.\egroup
  }{2017}]{refinenet}
Guosheng Lin, Anton Milan, Chunhua Shen, and Ian Reid.
\newblock Refinenet: Multi-path refinement networks for high-resolution
  semantic segmentation.
\newblock In {\em CVPR}, 2017.

\bibitem[\protect\citeauthoryear{Liu and Yin}{2019}]{fpenet}
Mengyu Liu and Hujun Yin.
\newblock Feature pyramid encoding network for real-time semantic segmentation.
\newblock {\em arXiv}, 2019.

\bibitem[\protect\citeauthoryear{Liu \bgroup \em et al.\egroup
  }{2017}]{liu2017learning}
Sifei Liu, Shalini De~Mello, Jinwei Gu, Guangyu Zhong, Ming-Hsuan Yang, and Jan
  Kautz.
\newblock Learning affinity via spatial propagation networks.
\newblock In {\em NIPS}, 2017.

\bibitem[\protect\citeauthoryear{Long \bgroup \em et al.\egroup }{2015}]{fcn}
Jonathan Long, Evan Shelhamer, and Trevor Darrell.
\newblock Fully convolutional networks for semantic segmentation.
\newblock In {\em CVPR}, 2015.

\bibitem[\protect\citeauthoryear{Mehta \bgroup \em et al.\egroup
  }{2019}]{espnetv2}
Sachin Mehta, Mohammad Rastegari, Linda Shapiro, and Hannaneh Hajishirzi.
\newblock Espnetv2: A light-weight, power efficient, and general purpose
  convolutional neural network.
\newblock In {\em Proceedings of the IEEE conference on computer vision and
  pattern recognition}, pages 9190--9200, 2019.

\bibitem[\protect\citeauthoryear{Mei \bgroup \em et al.\egroup }{2020}]{GDNet}
Haiyang Mei, Xin Yang, Yang Wang, Yuanyuan Liu, Shengfeng He, Qiang Zhang,
  Xiaopeng Wei, and Rynson~W.H. Lau.
\newblock Don't hit me! glass detection in real-world scenes.
\newblock In {\em Proceedings of the IEEE/CVF Conference on Computer Vision and
  Pattern Recognition (CVPR)}, June 2020.

\bibitem[\protect\citeauthoryear{Meinhardt \bgroup \em et al.\egroup
  }{2021}]{trackformer}
Tim Meinhardt, Alexander Kirillov, Laura Leal-Taixe, and Christoph
  Feichtenhofer.
\newblock Trackformer: Multi-object tracking with transformers.
\newblock {\em arXiv preprint arXiv:2101.02702}, 2021.

\bibitem[\protect\citeauthoryear{Poudel \bgroup \em et al.\egroup
  }{2018}]{contextnet}
Rudra~PK Poudel, Ujwal Bonde, Stephan Liwicki, and Christopher Zach.
\newblock Contextnet: Exploring context and detail for semantic segmentation in
  real-time.
\newblock {\em arXiv}, 2018.

\bibitem[\protect\citeauthoryear{Poudel \bgroup \em et al.\egroup
  }{2019}]{fastscnn}
Rudra~PK Poudel, Stephan Liwicki, and Roberto Cipolla.
\newblock Fast-scnn: fast semantic segmentation network.
\newblock {\em arXiv}, 2019.

\bibitem[\protect\citeauthoryear{Ronneberger \bgroup \em et al.\egroup
  }{2015}]{unet}
Olaf Ronneberger, Philipp Fischer, and Thomas Brox.
\newblock U-net: Convolutional networks for biomedical image segmentation.
\newblock In {\em MICCAI}, 2015.

\bibitem[\protect\citeauthoryear{Singh \bgroup \em et al.\egroup
  }{2018}]{multidata}
Ravinder Singh, Kuldeep~Singh Nagla, John Page, and John Page.
\newblock Multi-data sensor fusion framework to detect transparent object for
  the efficient mobile robot mapping.
\newblock {\em International Journal of Intelligent Unmanned Systems}, pages
  00--00, 2018.

\bibitem[\protect\citeauthoryear{Spataro \bgroup \em et al.\egroup
  }{2015}]{reachandgrasp}
R.~Spataro, R.~Sorbello, S.~Tramonte, G.~Tumminello, M.~Giardina, A.~Chella,
  and V.~La~Bella.
\newblock Reaching and grasping a glass of water by locked-in als patients
  through a bci-controlled humanoid robot.
\newblock {\em Frontiers in Human Neuroscience}, 357:e48--e49, 2015.

\bibitem[\protect\citeauthoryear{Sun \bgroup \em et al.\egroup
  }{2020}]{transtrack}
Peize Sun, Yi~Jiang, Rufeng Zhang, Enze Xie, Jinkun Cao, Xinting Hu, Tao Kong,
  Zehuan Yuan, Changhu Wang, and Ping Luo.
\newblock Transtrack: Multiple-object tracking with transformer.
\newblock {\em arXiv preprint arXiv:2012.15460}, 2020.

\bibitem[\protect\citeauthoryear{Vaswani \bgroup \em et al.\egroup
  }{2017}]{vaswani2017attention}
Ashish Vaswani, Noam Shazeer, Niki Parmar, Jakob Uszkoreit, Llion Jones,
  Aidan~N Gomez, {\L}ukasz Kaiser, and Illia Polosukhin.
\newblock Attention is all you need.
\newblock In {\em Advances in neural information processing systems}, pages
  5998--6008, 2017.

\bibitem[\protect\citeauthoryear{Wang \bgroup \em et al.\egroup
  }{2018}]{nonlocal}
Xiaolong Wang, Ross Girshick, Abhinav Gupta, and Kaiming He.
\newblock Non-local neural networks.
\newblock In {\em CVPR}, 2018.

\bibitem[\protect\citeauthoryear{Wang \bgroup \em et al.\egroup
  }{2019a}]{hrnet}
Jingdong Wang, Ke~Sun, Tianheng Cheng, Borui Jiang, Chaorui Deng, Yang Zhao,
  Dong Liu, Yadong Mu, Mingkui Tan, Xinggang Wang, et~al.
\newblock Deep high-resolution representation learning for visual recognition.
\newblock {\em arXiv}, 2019.

\bibitem[\protect\citeauthoryear{Wang \bgroup \em et al.\egroup
  }{2019b}]{lednet}
Yu~Wang, Quan Zhou, Jia Liu, Jian Xiong, Guangwei Gao, Xiaofu Wu, and
  Longin~Jan Latecki.
\newblock Lednet: A lightweight encoder-decoder network for real-time semantic
  segmentation.
\newblock In {\em ICIP}, 2019.

\bibitem[\protect\citeauthoryear{Wang \bgroup \em et al.\egroup
  }{2020}]{wang2020end}
Yuqing Wang, Zhaoliang Xu, Xinlong Wang, Chunhua Shen, Baoshan Cheng, Hao Shen,
  and Huaxia Xia.
\newblock End-to-end video instance segmentation with transformers.
\newblock {\em arXiv preprint arXiv:2011.14503}, 2020.

\bibitem[\protect\citeauthoryear{Xie \bgroup \em et al.\egroup
  }{2020}]{translab}
Enze Xie, Wenjia Wang, Wenhai Wang, Mingyu Ding, Chunhua Shen, and Ping Luo.
\newblock Segmenting transparent objects in the wild.
\newblock {\em arXiv preprint arXiv:2003.13948}, 2020.

\bibitem[\protect\citeauthoryear{Xu \bgroup \em et al.\egroup
  }{2015}]{transcut}
Yichao Xu, Hajime Nagahara, Atsushi Shimada, and Rin{-}ichiro Taniguchi.
\newblock Transcut: Transparent object segmentation from a light-field image.
\newblock In {\em ICCV}, 2015.

\bibitem[\protect\citeauthoryear{Yang \bgroup \em et al.\egroup
  }{2018}]{denseaspp}
Maoke Yang, Kun Yu, Chi Zhang, Zhiwei Li, and Kuiyuan Yang.
\newblock Denseaspp for semantic segmentation in street scenes.
\newblock In {\em CVPR}, 2018.

\bibitem[\protect\citeauthoryear{Yu \bgroup \em et al.\egroup }{2018}]{bisenet}
Changqian Yu, Jingbo Wang, Chao Peng, Changxin Gao, Gang Yu, and Nong Sang.
\newblock Bisenet: Bilateral segmentation network for real-time semantic
  segmentation.
\newblock In {\em ECCV}, 2018.

\bibitem[\protect\citeauthoryear{Yuan and Wang}{2018}]{ocnet}
Yuhui Yuan and Jingdong Wang.
\newblock Ocnet: Object context network for scene parsing.
\newblock {\em arXiv}, 2018.

\bibitem[\protect\citeauthoryear{Zhao \bgroup \em et al.\egroup
  }{2017}]{pspnet}
Hengshuang Zhao, Jianping Shi, Xiaojuan Qi, Xiaogang Wang, and Jiaya Jia.
\newblock Pyramid scene parsing network.
\newblock In {\em CVPR}, 2017.

\bibitem[\protect\citeauthoryear{Zhao \bgroup \em et al.\egroup }{2018}]{icnet}
Hengshuang Zhao, Xiaojuan Qi, Xiaoyong Shen, Jianping Shi, and Jiaya Jia.
\newblock Icnet for real-time semantic segmentation on high-resolution images.
\newblock In {\em ECCV}, 2018.

\bibitem[\protect\citeauthoryear{Zheng \bgroup \em et al.\egroup
  }{2015}]{zheng2015conditional}
Shuai Zheng, Sadeep Jayasumana, Bernardino Romera-Paredes, Vibhav Vineet,
  Zhizhong Su, Dalong Du, Chang Huang, and Philip~HS Torr.
\newblock Conditional random fields as recurrent neural networks.
\newblock In {\em ICCV}, 2015.

\bibitem[\protect\citeauthoryear{Zheng \bgroup \em et al.\egroup
  }{2020}]{zheng2020setr}
Sixiao Zheng, Jiachen Lu, Hengshuang Zhao, Xiatian Zhu, Zekun Luo, Yabiao Wang,
  Yanwei Fu, Jianfeng Feng, Tao Xiang, Philip~HS Torr, et~al.
\newblock Rethinking semantic segmentation from a sequence-to-sequence
  perspective with transformers.
\newblock {\em arXiv preprint arXiv:2012.15840}, 2020.

\bibitem[\protect\citeauthoryear{Zhou \bgroup \em et al.\egroup
  }{2017}]{zhou2017scene}
Bolei Zhou, Hang Zhao, Xavier Puig, Sanja Fidler, Adela Barriuso, and Antonio
  Torralba.
\newblock Scene parsing through ade20k dataset.
\newblock 2017.

\bibitem[\protect\citeauthoryear{Zhou \bgroup \em et al.\egroup
  }{2018}]{iros2018}
Zheming Zhou, Zhiqiang Sui, and Odest~Chadwicke Jenkins.
\newblock Plenoptic monte carlo object localization for robot grasping under
  layered translucency.
\newblock In {\em 2018 IEEE/RSJ International Conference on Intelligent Robots
  and Systems (IROS)}, pages 1--8. IEEE, 2018.

\bibitem[\protect\citeauthoryear{Zhu \bgroup \em et al.\egroup
  }{2020}]{deformdetr}
Xizhou Zhu, Weijie Su, Lewei Lu, Bin Li, Xiaogang Wang, and Jifeng Dai.
\newblock Deformable detr: Deformable transformers for end-to-end object
  detection.
\newblock {\em arXiv preprint arXiv:2010.04159}, 2020.

\end{thebibliography}
